\documentclass[11pt,a4paper]{article}
\usepackage[hyperref]{emnlp2020}
\usepackage{times}
\usepackage{latexsym}
\usepackage{tikz}
\usepackage{graphicx}

\usepackage{hyperref}
\usepackage{float}
\usepackage{footnote}
\usepackage{threeparttable}
\usepackage{enumitem}
\usepackage{bm}
\usepackage{subfig}
\usepackage{arydshln}
\usepackage{microtype}
\usepackage{multirow}

\usepackage{amsmath}
\usepackage{url}
\usepackage{graphicx}

\usepackage{booktabs}
\newenvironment{itemize*}%
 {\begin{itemize}%
  \setlength{\itemsep}{0pt}%
  \setlength{\parskip}{0pt}}%
 {\end{itemize}}
 \newenvironment{enumerate*}%
 {\begin{enumerate}%
  \setlength{\itemsep}{0pt}%
  \setlength{\parskip}{0pt}}%
 {\end{enumerate}}

\usepackage{lipsum}

\aclfinalcopy
\definecolor{myblue}{rgb}{0.9, 0.1, 0.94}
\definecolor{mygreen}{rgb}{0.64, 0.56, 0.88}
\definecolor{myyellow}{rgb}{0.98, 0.94, 0.75}
\definecolor{mygreen}{rgb}{0.68, 0.9, 0.6}

\definecolor{mot}{rgb}{0.961, 0.848, 0.902}
\definecolor{ori}{rgb}{0.8086, 0.9141, 0.7930}
\definecolor{sou}{rgb}{0.8398, 0.9219, 0.9688}
\definecolor{sub}{rgb}{0.9766, 0.8906, 0.8008}
\definecolor{rep}{rgb}{0.738, 0.762, 0.781}
\definecolor{cla}{rgb}{0.9727, 0.9531, 0.7617}
\definecolor{cmp}{rgb}{0.8398, 0.9219, 0.9688}
\definecolor{sum}{rgb}{0.8398, 0.8438, 0.9805}
\definecolor{po}{rgb}{0.96, 0.80, 0.88}
\definecolor{pm}{rgb}{0.80, 0.90, 0.95}

\definecolor{neg}{rgb}{0.924, 0.320, 0.313}
\definecolor{pos}{rgb}{0.328, 0.449, 0.823}

\definecolor{weizhey}{rgb}{0.43, 0.71, 0.40}

\newcommand{\mycbox}[1]{\tikz{\path[draw=#1,fill=#1] (0,0) rectangle (0.25cm,0.25cm);}}

\title{Can We Automate Scientific Reviewing?}

\author{Weizhe Yuan \\
  Carnegie Mellon University \\
  \texttt{weizhey@cs.cmu.edu} \\
  \And
  Pengfei Liu \thanks{\ \  Corresponding author.} \\
  Carnegie Mellon University \\
\texttt{pliu3@cs.cmu.edu}
  \And
  Graham Neubig \\
  Carnegie Mellon University \\
\texttt{gneubig@cs.cmu.edu}
  }

\date{}

\begin{document}
\maketitle

\begin{center}\normalsize\textbf{TL;QR}\end{center}

\begin{quotation}
This paper proposes to use NLP models to generate reviews for scientific papers . The model is trained on the ASAP-Review dataset and evaluated on a set of metrics to evaluate the quality of the generated reviews . It is found that the model is not very good at summarizing the paper , but it is able to generate more detailed reviews that cover more aspects of the paper than those created by humans . The paper also finds that both human and automatic reviewers exhibit varying degrees of bias and biases , and that the system generate more biased reviews than human reviewers.(\textcolor{gray}{``Too Long; Quick Read'', this paragraph, is generated by our system.})
\end{quotation}


\begin{abstract}

The rapid development of science and technology has been accompanied by an exponential growth in peer-reviewed scientific publications.
At the same time, the review of each paper is a laborious process that must be carried out by subject matter experts.
Thus, providing high-quality reviews of this growing number of papers is a significant challenge.
In this work, we ask the question ``\textit{can we automate scientific reviewing?}'', discussing the possibility of using state-of-the-art natural language processing (NLP) models to generate first-pass peer reviews for scientific papers.
Arguably the most difficult part of this is defining what a ``good'' review is in the first place, so we first discuss possible evaluation measures for such reviews.
We then collect a dataset of papers in the machine learning domain, annotate them with different aspects of content covered in each review, and train targeted summarization models that take in papers to generate reviews.
Comprehensive experimental results show that system-generated reviews tend to touch upon more aspects of the paper than human-written reviews, but the generated text can suffer from lower constructiveness for all aspects except the explanation of the core ideas of the papers, which are largely factually correct.
We finally summarize \emph{eight} challenges in the pursuit of a good review generation system together with potential solutions, which, hopefully, will inspire more future research on this subject.
We make all code, and the dataset publicly available:
\url{https://github.com/neulab/ReviewAdvisor}
as well as a \textit{ReviewAdvisor} system:
\url{http://review.nlpedia.ai/} (See demo screenshot in \ref{sec:screenshot}).
\textcolor{gray}{The review of this paper (without TL;QR section) written by the system of this paper can be found \ref{sec:self-review}}
\end{abstract}

\section{Introduction}
The number of published papers is growing exponentially
\cite{tabah1999literature,de2009bibliometrics,bornmann2015growth}.
While this may be positively viewed as indicating acceleration of scientific progress, it also poses great challenges for researchers, both in reading and synthesizing the relevant literature for one’s own benefit, and for performing \emph{peer review} of papers to vet their correctness and merit.
With respect to the former, a large body of existing work explores automatic summarization of a paper or a set of papers for automatic survey generation~\cite{mohammad-etal-2009-using,jha-etal-2013-system,jha-etal-2015-content,DBLP:conf/aaai/JhaCR15,DBLP:conf/aaai/YasunagaKZFLFR19,cohan-etal-2018-discourse,xing-etal-2020-automatic}.
However, despite the fact that peer review is an important, but laborious part of our scientific process, automatic systems to aid in the peer review process remain relatively underexplored.
\citet{bartoli2016your} investigated the feasibility of generating reviews by surface-level term replacement and sentence reordering, and \citet{wang2020reviewrobot} (contemporaneously and independently) propose a two-stage information extraction and summarization pipeline to generate paper reviews.
However, both do not extensively evaluate the quality or features of the generated review text.

In this work, we are concerned with providing at least a preliminary answer to the ambitious over-arching question: \textit{can we automate scientific reviewing?}
Given the complexity of understanding and assessing the merit of scientific contributions,
we do not expect an automated system to be able to match a well-qualified and meticulous human reviewer at this task any time soon.
However, some degree of review automation may assist reviewers in their assessments, or provide guidance to junior reviewers who are just learning the ropes of the reviewing process.
Towards this goal, we examine two concrete research questions, the answers to which are prerequisites to building a functioning review assistant:

\noindent \textbf{Q1: What are the desiderata of a good automatic reviewing system, and how can we quantify them for evaluation?}
Before developing an automatic review system, we first must quantify what constitutes a good review in the first place.
The challenge of answering this question is that a review commonly involves both objective (e.g.~``lack of details necessary to replicate the experimental protocol'') and subjective aspects (e.g.~``lack of potential impact'').
Due to this subjectivity, defining a ``good'' review is itself somewhat subjective.

As a step towards tackling this challenge, we argue that it is possible to view review generation as a task of \emph{aspect-based scientific paper summarization}, where the summary not only tries to summarize the core idea of a paper, but also assesses specific aspects of that paper (e.g.~novelty or potential impact).
We evaluate review quality from multiple perspectives, in which we claim a good review not only should make a good summary of a paper but also consist of factually correct and fair comments from diverse aspects, together with informative evidence.

To operationalize these concepts, we build a dataset of reviews, named \texttt{ASAP-Review}\footnote{\texttt{AS}pect-enh\texttt{A}nced \texttt{P}eer Review dataset} from machine learning domain, and make fine-grained annotations of aspect information for each review, which provides the possibility for a richer evaluation of generated reviews.

\noindent \textbf{Q2: Using state-of-the-art NLP models, to what extent can we realize these desiderata?}
We provide an initial answer to this question by using the aforementioned dataset to train state-of-the-art summarization models to generate reviews from scientific papers, and evaluate the output according to our evaluation metrics described above.
We propose different architectural designs for this model, which we dub \textit{ReviewAdvisor} (\S\ref{sec:model}), and comprehensively evaluate them, interpreting their relative advantages.

\noindent \textbf{Lastly, we highlight our main observations and conclusions}:

\noindent \textit{(1) What are review generation systems (not) good at?}
Most importantly, we find the constructed automatic review system \emph{generates non-factual statements} regarding many aspects of the paper assessment, which is a serious flaw in a high-stakes setting such as reviewing.
However, there are some bright points as well.
For example, it \emph{can often precisely summarize the core idea} of the input paper, which can be either used as a draft for human reviewers or help them (or general readers) quickly understand the main idea of the paper to be reviewed (or pre-print papers).
It can also generate reviews that \emph{cover more aspects} of the paper's quality than those created by humans, and provide evidence sentences from the paper.
These could potentially provide a preliminary template for reviewers and help them quickly identify salient information in making their assessment.

\noindent \textit{(2) Will the system generate biased reviews?}
Yes. We present methods to identify and quantify potential biases in reviews (\S\ref{sec:bias-analysis}), and find that both human and automatic reviewers exhibit varying degrees of bias.
(i) regarding native vs. non-native English speakers: papers of native English speakers tend to obtain higher scores on ``\texttt{Clarity}'' from human reviewers than non-native English ones,\footnote{Whether this actually qualifies as ``bias'' is perhaps arguable. Papers written by native English speakers may be more clear due to lack of confusing grammatical errors, but the paper may actually be perfectly clear but give the impression of not being clear because of grammatical errors.} but the automatic review generators narrow this gap.
Additionally, system reviewers are harsher than human reviewers when commenting regarding the paper's ``\texttt{Originality}'' for non-native English speakers.
(ii)
regarding anonymous vs. non-anonymous submissions:
 both human reviewers and system reviewers favor non-anonymous papers, which have been posted on non-blind preprint servers such as arXiv\footnote{\url{https://arxiv.org/}} before the review period,  more than anonymous papers in all aspects.

Based on above mentioned issues, we claim that \textbf{a review generation system can not replace human reviewers at this time, instead, it may be helpful as part of a machine-assisted human review process}.
Our research also enlightens what's next in pursuing a better method for automatic review generation or assistance and we summarize \emph{eight challenges}  that can be explored for future directions in \S\ref{sec:challenges}.

\section{What Makes a Good Peer Review?}
Although peer review has been adopted by most journals and conferences to identify important and relevant research, its effectiveness is being continuously questioned \citep{Smith2006PeerRA, langford2015arbitrariness, tomkins2017reviewer, gao-etal-2019-rebuttal, rogers-augenstein-2020-improve}.

As concluded by \citet{jefferson2002measuring}: ``\textit{Until we have properly defined the objectives of peer-review, it will remain almost impossible to assess or improve its effectiveness.}''
Therefore 
we first discuss the possible objectives of peer review.

\subsection{Peer Review for Scientific Research}
\label{sec:eval-definition}
A research paper is commonly first reviewed by several committee members who usually assign one or several \textit{scores} and give detailed comments.
The comments, and sometimes scores, cover diverse \textit{aspects} of the paper (e.g. ``{clarity},'' ``potential impact''; detailed in \S\ref{sec:aspect_def}), and these aspects are often directly mentioned in review forms of scientific conferences or journals.%
\footnote{For example, one example from ACL can be found at: \url{https://acl2018.org/downloads/acl\_2018\_review\_form.html}}

Then a senior reviewer  will often make a \textit{final decision} (i.e., ``reject'' or ``accept'') and provide comments summarizing the decision (i.e., a \textit{meta-review}).

After going through many review guidelines\footnote{\url{https://icml.cc/Conferences/2020/ReviewerGuidelines} \url{https://NeurIPS.cc/Conferences/2020/PaperInformation/ReviewerGuidelines}, \url{https://iclr.cc/Conferences/2021/ReviewerGuide}} and resources about how to write a good review%
\footnote{\url{https://players.brightcove.net/3806881048001/rFXiCa5uY_default/index.html?videoId=4518165477001}, \url{https://soundcloud.com/nlp-highlights/77-on-writing-quality-peer-reviews-with-noah-a-smith}, \url{https://www.aclweb.org/anthology/2020.acl-tutorials.4.pdf}, \url{https://2020.emnlp.org/blog/2020-05-17-write-good-reviews}}
we summarize \emph{some} of the most frequently mentioned desiderata below:
\begin{enumerate}

    \item \textbf{Decisiveness:} A good review should take a clear stance, selecting high-quality submissions for publication and suggesting others not be accepted \citep{jefferson2002effects, Smith2006PeerRA}.
    \item \textbf{Comprehensiveness:} A good review should be well-organized, typically starting with a brief summary of the paper's contributions, then following with opinions gauging the quality of a paper from different aspects. Many review forms explicitly require evaluation of different aspects to encourage comprehensiveness.
    \item \textbf{Justification:} A good review should provide specific reasons for its assessment, particularly whenever it states that the paper is lacking in some aspect.
    This justification also makes the review more constructive (another oft-cited desiderata of reviews), as these justifications provide hints about how the authors could improve problematic aspects in the paper \citep{xiong-litman-2011-automatically}.
    \item \textbf{Accuracy:} A review should be factually correct, with the statements contained therein not being demonstrably false.
    \item \textbf{Kindness:} A good review should be kind and polite in language use.
\end{enumerate}

Based on above desiderata, we make a first step towards evaluation of reviews for scientific papers and characterize a ``good'' review from multiple perspectives.

\subsection{Multi-Perspective Evaluation}
\label{sec: multi-pers-eval}

\begin{table}[t]
    \footnotesize
    \renewcommand{\arraystretch}{1.3}
  \centering
    \begin{tabular}{llrc}
    \toprule
    \textbf{Desiderata} & \textbf{Metrics} & \multicolumn{1}{l}{\textbf{Range}} & \multicolumn{1}{l}{\textbf{Automated}} \\
    \midrule
    Decisiveness & \textsc{RAcc}  &   [-1, 1]    & No \\ \hdashline
    \multicolumn{1}{c}{\multirow{2}[0]{*}{Comprehen.}} & \textsc{ACov}  &   [0, 1]    & Yes \\
          & \textsc{ARec}  &  [0, 1]     & Yes \\ \hdashline
    Justification & \textsc{Info}  &    [0, 1]   & No \\ \hdashline
    \multicolumn{1}{l}{\multirow{2}[1]{*}{Accuracy}} & \textsc{SAcc}  &   [0, 1]    & No \\
          & \textsc{ACon}  &   [0, 1]    & No \\
    \hdashline
    \multirow{2}[2]{*}{Others} & ROUGE &  [0, 1]     & Yes \\
          & BERTScore &  [-1, 1]     & Yes \\
    \bottomrule
    \end{tabular}
    \caption{\label{tab:multi-metric}Evaluation metrics from different perspectives. ``Range'' represents the range value of each metric. ``Automated'' denotes if metrics can be obtained automatically.}
\end{table}%

Given input paper $D$ and meta-review $R^m$, our goal is to evaluate the quality of review $R$, which can be either manually or automatically generated.
We also introduce a function $\textsc{Dec}(D) \in \{1,-1\}$ that indicates the final decision of a given paper reached by the meta-review: ``\texttt{accept}'' or ``\texttt{reject}''.
Further, $\textsc{Rec}(R) \in \{1,0,-1\}$ represents the acceptance recommendation of a particular review: ``\texttt{accept},'' ``\texttt{neutral},'' or ``\texttt{reject} (see Appendix~\ref{app:detail-metric} for details).

Below, we discuss evaluation metrics that can be used to approximate the desiderata of reviews described in the previous section. And we have summarized them in Tab.~\ref{tab:multi-metric}.

\subsubsection{D1: Decisiveness}
First, we tackle the \emph{decisiveness}, as well as accuracy of the review's recommendation, through \textbf{Recommendation Accuracy (\textsc{RAcc})}.
Here we use the final decision regarding a paper and measure whether the acceptance implied by the review $R$ is consistent with the actual accept/reject decision of the reviewed paper.
It is calculated as:
\begin{equation}
\label{eq:sfac}
    \text{RAcc}(R) = \textsc{Dec}(D)\times\textsc{Rec}(R)
\end{equation}

A higher score indicates that the review more decisively and accurately makes an acceptance recommendation.

\subsubsection{D2: Comprehensiveness}

A comprehensive review should touch on the quality of different aspects of the paper, which we measure using  a metric dubbed \textbf{Aspect Coverage (\textsc{ACov})}. Specifically,
given a review $R$, aspect coverage measures how many aspects (e.g. \texttt{clarity}) in a predefined aspect typology (in our case, \S\ref{sec:aspect_def}) have been covered by $R$.

In addition, we propose another metric \textbf{Aspect Recall (\textsc{ARec})}, which explicitly takes the meta-review $R^m$ into account.
Because the meta-review is an authoritative summary of all the reviews for a paper, it provides an approximation of which aspects, and with which sentiment polarity, should be covered in a review.
Aspect recall counts how many aspects in meta-review $R^m$ are covered by general review $R$, with higher aspect recall indicating better agreement with the meta-review.%
\footnote{
Notably, this metric potentially biases towards high scores for reviews that were considered in the writing of the meta-review.
Therefore, higher aspect recall is not the only goal, and should be taken together with other evaluation metrics.
}

\subsubsection{D3: Justification}

As defined in \S\ref{sec:eval-definition}, a good peer review should provide hints about how the author could improve problematic aspects. For example,  when reviewers comment: ``this paper lacks important references'', they should also list these relevant works.
To satisfy this justification desideratum, we define a metric called \textbf{Informativeness (\textsc{Info})} to quantify how many negative comments\footnote{We only consider whether the reviewer has provided enough evidence for negative opinions since we find that most human reviewers rarely provide evidence for their positive comments.} are accompanied by corresponding evidence.

First, let $n_{\text{na}}(R)$ denote the number of aspects in $R$ with negative sentiment polarity. $n_{\text{nae}}(R)$ denotes the number of aspects with negative sentiment polarity that are supported by evidence. The judgement of supporting evidence is conducted manually (details in Appendix \ref{app:detail-metric}). \textsc{Info} is calculated as:

\begin{equation}
\label{eq:info}
    \text{Info}(R) = \frac{n_{\text{nae}}(R)}{n_{\text{na}}(R)}
\end{equation}

And we set it to be 1 when there are no negative aspects mentioned in a review.

\subsection{D4: Accuracy}
We use two measures to evaluate the accuracy of assessments.
First, we use \textbf{Summary Accuracy (\textsc{SAcc})} to measure how well a review summarizes contributions of a paper. It takes value of 0, 0.5, or 1, which evaluates the summary part of the review as incorrect/absent, partially correct, and correct. The correctness judgement is performed manually, with details listed in Appendix \ref{app:detail-metric}.

\textsc{Info} implicitly requires that negative aspects should be supported with evidence, ignoring the quality of this evidence. However, to truly help to improve the quality of a paper, the evidence for negative aspects should be factual as well. Here we propose \textbf{Aspect-level Constructiveness (\textsc{ACon})}, the percentage of the supporting statements $n_{\text{nae}}(R)$ that are judged as valid support by human annotators.
If $n_{\text{nae}}(R)$ is 0, we set its \textsc{ACon} as 1.
This metric will implicitly favor reviews that do not provide enough evidence for negative aspects. However, in this case, the \textsc{Info} of those reviews will be rather low.
The details of evaluating ``validity'' are also described in Appendix~\ref{app:detail-metric}.

\subsection{D5: Kindness}
While kindness is very important in maintaining a positive research community, accurately measuring it computationally in a nuanced setting such as peer review is non-trivial.
Thus, we leave the capturing of kindness in evaluation to future work.

\subsection{ Similarity to Human Reviews}
For automatically generated reviews, we also use \textbf{Semantic Equivalence} metrics to measure the similarity between generated reviews and reference reviews.
The intuition is that while human reviewers are certainly not perfect, knowing how close our generated reviews are to existing human experts may be informative. Here, we investigate two specific metrics: ROUGE \cite{lin2003automatic} and BERTScore \cite{zhang2019bertscore}. The former measures the surface-level word match while the latter measures the distance in embedding space. Notably, for each source input, there are multiple reference reviews. When aggregating ROUGE and BERTScore, we take the maximum instead of average since it is not necessary for generated reviews to be close to all references.

\section{Dataset}
\label{sec: dataset}
Next, in this section we introduce how we construct a review dataset with more fine-grained metadata, which can be used for system training and the multiple perspective evaluation of reviews.

\subsection{Data Collection}
\begin{figure*}
    \centering
    \includegraphics[width=0.92\linewidth]{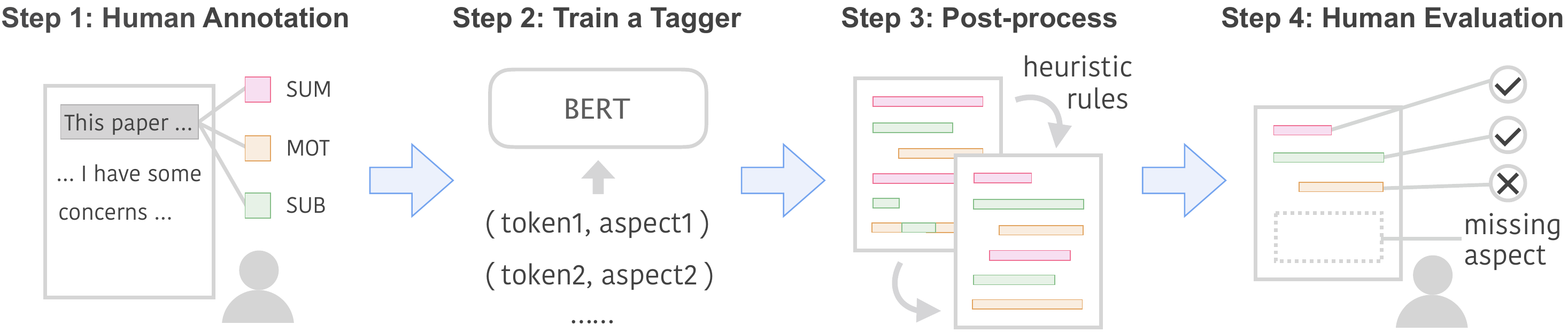}
    \caption{Data annotation pipeline.}
    \label{fig:pipeline}
\end{figure*}
The advent of the Open Peer Review system\footnote{\url{https://openreview.net/}} makes it possible to access review data for analysis or model training/testing.
One previous work \cite{kang18naacl} attempts to collect reviews from several prestigious publication venues including the Conference of the Association of Computational Linguistics (ACL) and the International Conference on Learning Representations (ICLR). However, there were not nearly as many reviews accumulated in OpenReview at that time\footnote{During that time, there are no reviews of ICLR from 2018 to 2020 nor reviews of NeurIPS from 2018 to 2019.} and other private reviews only accounted for a few hundred.
Therefore we decided to collect our own dataset \texttt{As}pect-enh\texttt{a}nced \texttt{P}eer Review (\texttt{ASAP-Review}).

We crawled ICLR papers from 2017-2020 through OpenReview\footnote{\url{https://openreview.net}} and NeurIPS papers from 2016-2019 through NeurIPS Proceedings.\footnote{\url{http://papers.NeurIPS.cc}}
For each paper's review, we keep as much metadata information as possible.
Specifically, for each paper, we include following metadata information that we can obtain from the review web page:
\begin{itemize*}
    \item \textit{Reference reviews}, which are written by a committee member.
    \item \textit{Meta reviews}, which are commonly written by an area chair (senior committee member).
    \item \textit{Decision}, which denotes a paper's final ``accept'' or ``reject'' decision.
    \item \textit{Other} information like url, title, author, etc.
\end{itemize*}

We used Allenai Science-parse\footnote{\url{https://github.com/allenai/science-parse}} to parse the pdf of each paper and keep the structured textual information (e.g., titles, authors, section content and references). The basic statistics of our \texttt{ASAP-Review} dataset is shown in Tab.~\ref{table1}.

\begin{table}[h]
\centering
\begin{threeparttable}
\small
    \begin{tabular*}{0.48\textwidth}{@{}lrrr@{}}
\toprule
                            & ICLR & NeurIPS & Both \\ \midrule
\hspace{3mm}Accept                      & 1,859 & 3,685 & 5,544 \\
\hspace{3mm}Reject                      & 3,333 & 0    & 3,333 \\
\hspace{3mm}Total                       & 5,192 & 3,685 & 8877 \\
\hspace{3mm}Avg. Full Text Length       & 7,398 & 5,916 & 6782 \\
\hspace{3mm}Avg. Review Length          & 445  & 411  & 430 \\
\hspace{3mm}\# of Reviews               & 15,728 & 12,391 & 28,119 \\
\hspace{3mm}\# of Reviews per Paper     &  3.03   &  3.36 & 3.17 \\ \bottomrule
\end{tabular*}
    \caption{Basic statistics of \texttt{ASAP-Review} dataset. Note that NeurIPS only provide reviews for accepted papers to the public.}
    \label{table1}
\end{threeparttable}
\end{table}

\subsection{Aspect-enhanced Review Dataset}
\label{sec:aspect-enhanced}
Although reviews exhibit internal structure, for example, as shown in Fig.~\ref{fig:multi-view}, reviews commonly start with a paper summary, followed by different aspects of opinions, together with evidence. In practice, this useful structural information cannot be obtained directly.
Considering that fine-grained information about the various aspects touched on by the review plays an essential role in review evaluation, we conduct aspect annotation of those reviews.
To this end, we first (i) introducing an aspect typology and (ii) perform human annotation.

\subsubsection{Aspect Typology and Polarity}
\label{sec:aspect_def}
We define a typology that contains 8 aspects, which follows the ACL review guidelines\footnote{\url{https://acl2018.org/downloads/acl_2018_review_form.html}. We manually inspected several review guidelines from ML conferenecs and found the typology in ACL review guideline both general and comprehensive.}
with small modifications, which are \textit{Summary} (SUM), \textit{Motivation/Impact} (MOT)
, \textit{Originality} (ORI), \textit{Soundness/Correctness} (SOU), \textit{Substance} (SUB), \textit{Replicability} (REP), \textit{Meaningful Comparison} (CMP) and \textit{Clarity} (CLA). The detailed elaborations of each aspect can be found in Supplemental Material \ref{supp:data-annotation}. Inside the parentheses are what we will refer to each aspect for brevity.
To take into account whether the comments regarding each aspect are positive or negative, we also mark whether the comment is positive or negative for every aspect (except summary).

\subsubsection{Aspect Annotation}
\label{sec:data-annotate}
Overall, the data annotation involves four steps that are shown in Fig.~\ref{fig:pipeline}.

\begin{figure}[htbp]
    \centering
    \includegraphics[width=1\linewidth]{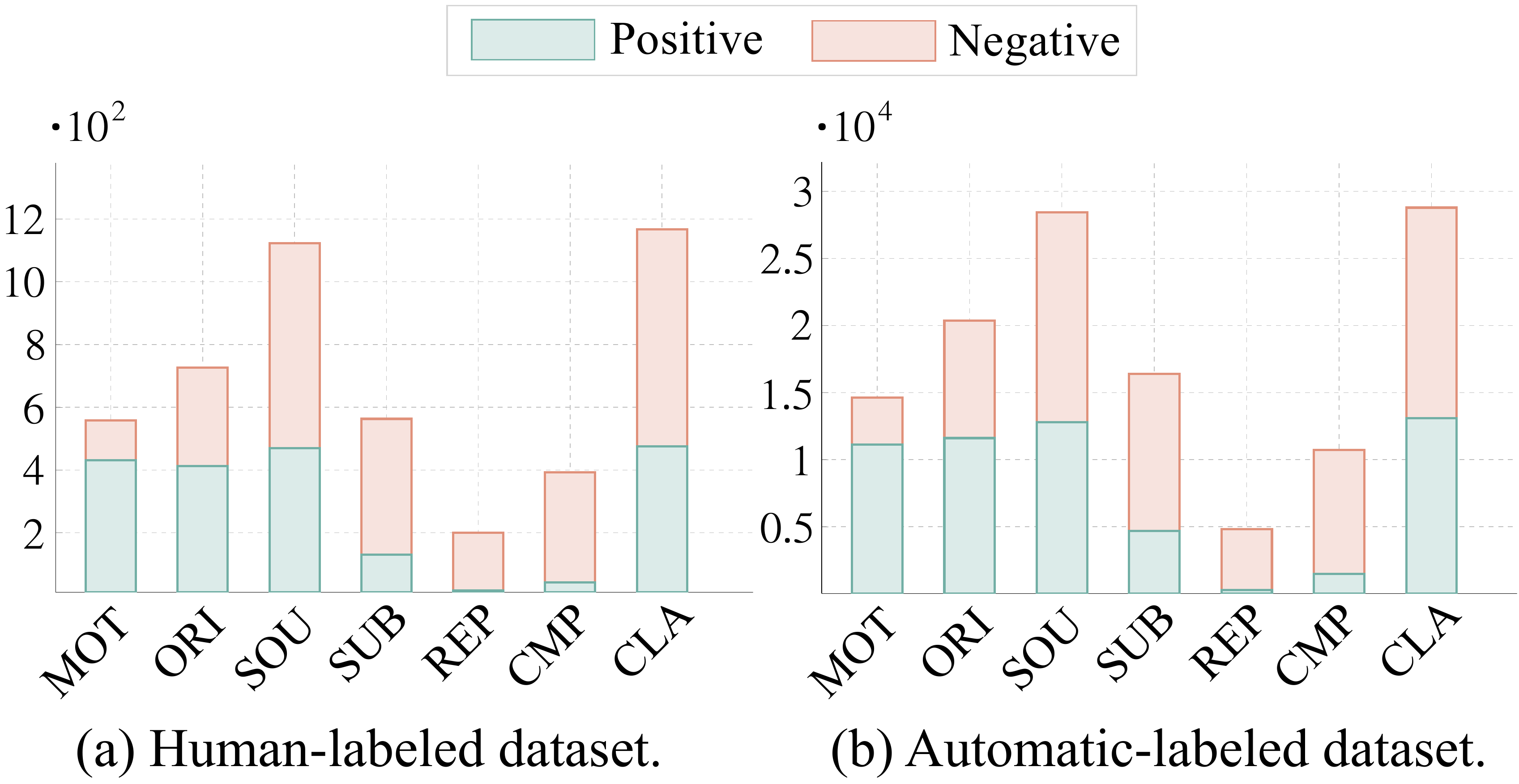}
    \caption{(a) and (b) represent distributions over seven aspects obtained by human and BERT-based tagger respectively. Red bins represent positive sentiment while green ones suggest negative sentiment. We omit ``Sum'' aspect since there is no polarity  definition of it.}
    \label{fig:statistic}
\end{figure}

\paragraph{Step 1: Manual Annotation}
To manually annotate aspects in reviews, we first set up a data annotation platform using Doccano.\footnote{\url{https://github.com/doccano/doccano}}
We asked 6 students from ML/NLP backgrounds to annotate the dataset. We asked them to tag an appropriate text span that indicates a specific aspect.
For example, ``\setlength{\fboxsep}{2pt}\colorbox{po}{The results are new}$_\text{[Positive Originality]}$ \setlength{\fboxsep}{1pt}\colorbox{pm}{and important to this field}$_\text{[Positive Motivation]}$''. The detailed annotation guideline can be found in Supplemental Material \ref{supp:data-annotation}.
Each review is annotated by two annotators and the lowest pair-wise Cohen kappa is 0.653, which stands for substantial agreement. In the end, we obtained 1,000 human-annotated reviews in total. The aspect statistics in this dataset are shown in Fig.~\ref{fig:statistic}-(a).

\paragraph{Step 2: Training an Aspect Tagger}
Since there are over 20,000 reviews in our dataset, using human labor to annotate them all is unrealistic. Therefore, we use the annotated data we do have to train an aspect tagger and use it to annotate the remaining reviews.
The basic architecture of our aspect tagger contains a pre-trained model BERT \cite{devlin2019bert} and a multi-layer perceptron. The training details can be found in Appendix \ref{app:train-tagger}.

\paragraph{Step 3: Post-processing}
However, after inspecting the automatically labeled dataset, we found that there appears to be some common problems such as interleaving different aspects and inappropriate boundaries. To address those problems,  we used seven heuristic rules to refine the prediction results and they were executed sequentially. The detailed heuristics can be found in Appendix \ref{sec:heuristics}. An example of our model prediction after applying heuristic rules is shown in Appendix \ref{sec:example-annotate}. Fig.~\ref{fig:statistic}-(b) shows the distribution of all reviews over different aspects.
As can be seen, the relative number of different aspects and the ratio of positive to negative are very similar across human and automatic annotation.

\paragraph{Step 4: Human Evaluation}
To evaluate the data quality of reviews' aspects, we conduct human evaluation. Specifically, we measure both aspect precision and aspect recall for our defined 15 aspects.

We randomly chose 300 samples from our automatically annotated dataset and  assigned each sample to three different annotators to judge the annotation quality. As before, these annotators are all from ML/NLP backgrounds.

The detailed calculation for aspect precision and aspect recall can be found in Appendix~\ref{app:asp-pre-and-asp-recall}.
Under these criteria, we achieved $92.75\%$ aspect precision and 85.19\% aspect recall. The fine-grained aspect precision and aspect recall for each aspect is shown in Tab.~\ref{tab:asp-cov-rec}. The aspect recall for positive replicability is low. This is due to the fact that there are very few mentions of positive replicability. And in our human evaluation case, the system identified one out of two, which results in 50\%. Other than that, the precision and recall are much higher.%
\footnote{
The recall numbers for negative aspects are lower than positive aspects. However, we argue that this will not affect the fidelity of our analysis much because (i) we observe that the imperfect recall is mostly (over $85\%$) caused by partial recognition of the same negative aspect in a review instead of inability to recognize at least one. This will not affect our calculation of \textit{Aspect Coverage} and \textit{Aspect Recall} very much. (ii) The imperfect recall will slightly pull up \textit{Aspect Score} (will discuss in \S\ref{sec:measure_bias}), but the trend will remain the same.
}

Besides, one thing to mention is that our evaluation criterion is very strict, and it thus acts as a lower bound for these two metrics.

\begin{table}[t]
    \footnotesize
    \renewcommand{\arraystretch}{1.3}
    \centering
    \setlength\tabcolsep{8pt}
    \begin{tabular*}{0.45\textwidth}{lcrr}
    \toprule
        \textbf{Aspect} & \textbf{Polarity} & \textbf{Precision} & \textbf{Recall} \\ \hline
        Summary &  & 95\% & 100\% \\ \hdashline
        \multicolumn{1}{l}{\multirow{2}[0]{*}{Motivation}} & + & 94\% & 89\% \\
         & -- & 72\% & 71\% \\ \hdashline
        \multicolumn{1}{l}{\multirow{2}[0]{*}{Originality}} & + & 95\% & 87\% \\
         & -- & 94\% & 80\% \\ \hdashline
        \multicolumn{1}{l}{\multirow{2}[0]{*}{Soundness}} & + & 95\% & 98\% \\
         & -- & 92\% & 79\% \\ \hdashline
        \multicolumn{1}{l}{\multirow{2}[0]{*}{Substance}} & + & 90\% & 94\% \\
         & -- & 90\% & 78\% \\ \hdashline
        \multicolumn{1}{l}{\multirow{2}[0]{*}{Replicability}} & + & 100\% & 50\% \\
         & -- & 77\% & 71\% \\ \hdashline
        \multicolumn{1}{l}{\multirow{2}[0]{*}{Clarity}} & + & 97\% & 92\% \\
         & -- & 92\% & 73\% \\ \hdashline
        \multicolumn{1}{l}{\multirow{2}[0]{*}{Comparison}} & + & 85\% & 100\% \\
         & -- & 94\% & 94\% \\ \bottomrule
    \end{tabular*}
    \caption{\label{tab:asp-cov-rec}Fine-grained aspect precision and recall for each aspect. + denotes positive and -- denotes negative.
    }
\end{table}

\section{Scientific Review Generation}
\label{sec:model}

\subsection{Task Formulation}

The task of scientific review generation can be conceptualized an \textit{aspect-based scientific paper summarization} task but with a few important differences.
Specifically,  most current works summarize a paper
(i) either from an ``\textit{author view}'' that only use content written by the author to form a summary \cite{Cohan2018ADA, Xiao2019ExtractiveSO, Erera2019ASS,Cohan2018ADA,Cachola2020TLDRES},
(ii) or from a ``\textit{reader view}'' that argues a paper's summary should take into account the view of those in the research community \cite{10.5555/1599081.1599168, cohan2017scientific, Yasunaga2019ScisummNetAL}.

\begin{figure}[t]
    \centering
    \includegraphics[width=0.47\textwidth]{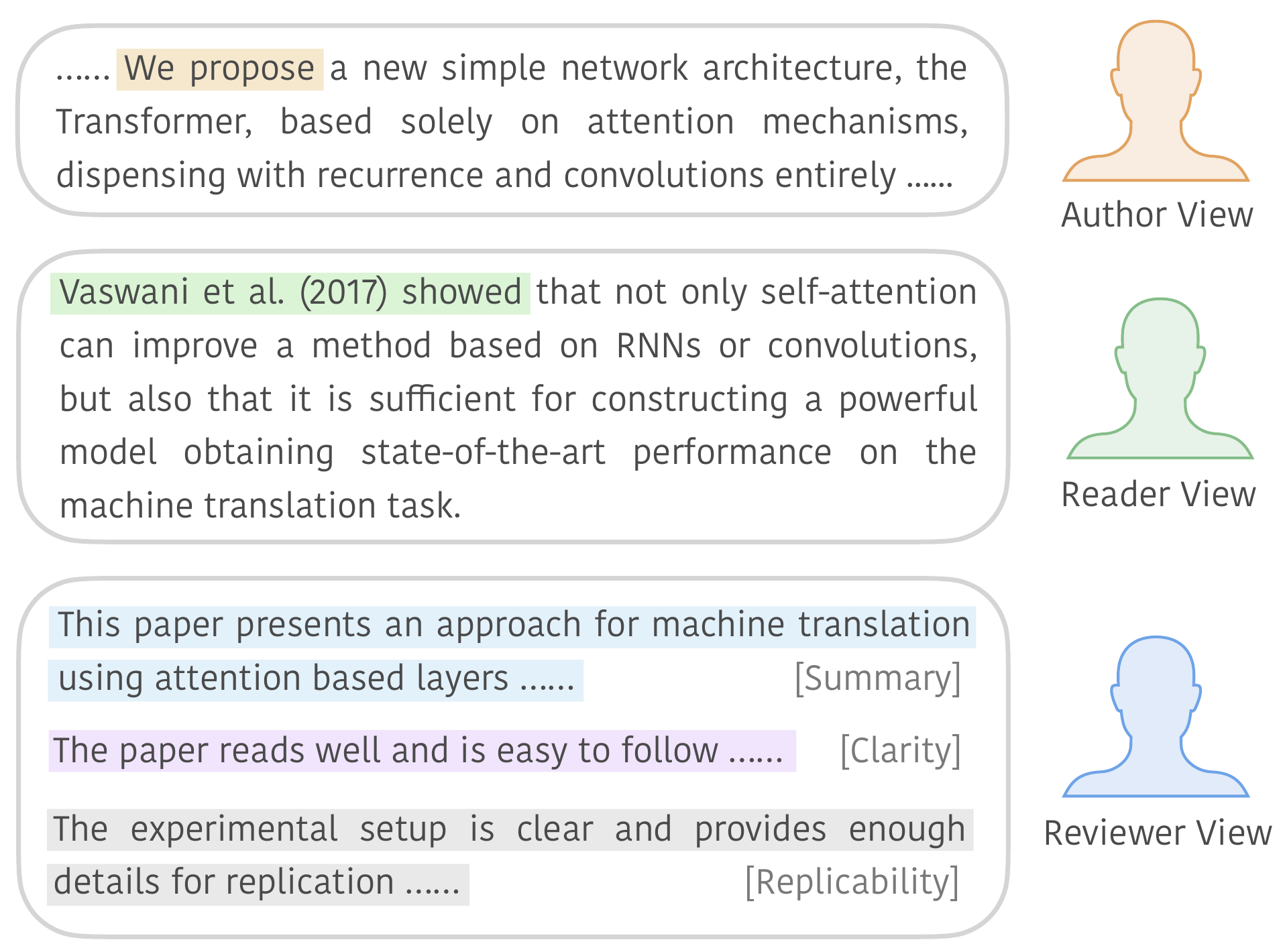}
    \caption{Summarization from three different views for the paper ``Attention Is All You Need" \citep{vaswani2017attention}. Summareis from three views (author, reader, reviewer) comes from the paper's abstract, citance (i.e., a paper that cites this paper) and peer review respectively.}
    \label{fig:multi-view}
\end{figure}

In this work, we extend the view of scientific paper summarization from ``author'' or ``reader'' to ``reviewer'', and claim that a good summary of a scientific paper can not only reflect the core idea but also contains critical comments from different aspects made by domain experts, which usually requires knowledge beyond the source paper itself.
The advantages lie in:
(i) \textit{authors}: helping them identify weak points in their paper and make it stronger.
(ii) \textit{reviewers}: relieving them from some of the burden of reviewing process.
(iii) \textit{readers}: helping them quickly grasp the main idea of the paper and letting them know what ``domain experts'' (our system) comments on the paper are.
The three views of scientific paper summarization are shown in Fig.~\ref{fig:multi-view}.

\subsection{System Design}
\label{sec:sysdesign}
Despite the fact that our dataset contains fewer training samples compared with other benchmark summarization datasets,
the few-shot learning ability of recent contextualized pre-trained models \cite{radford2019language,brown2020language,cachola-etal-2020-tldr}
still put training a passable review generation system from this dataset within grasp.
We use BART \citep{Lewis2019BARTDS}, which is a denoising autoencoder for pretraining sequence-to-sequence models, as our pre-trained model since it has shown superior performance on multiple generation tasks.

However, even if we can take the advantage of this pre-trained model, how to deal with lengthy text in the context of using a pre-trained model  (BART, for example, has a standard length limit of 1024
since it was pre-trained on texts of this size)
remains challenging.
After multiple trials, we opted for a two-stage method detailed below, and describe other explorations that were less effective in Appendix~\ref{app: bart4long}.

\subsubsection{Two-stage Systems for Long Documents}
\label{sec:2stage}

Instead of regarding text generation as a holistic process, we decompose it into two steps, using an \textit{extract-then-generate} paradigm \citep{chen2018fast, gehrmann2018bottom, subramanian2019extractive,dou2020gsum}.
Specifically, we first perform content selection, extracting salient text pieces from source documents (papers), then generate summaries based on these extracted texts.

To search for an effective way to select content that is most useful for constructing a review generation system, we operationalize the first extraction step in several ways.
One thing to notice is that the extraction methods we use here mainly focus on heuristics.
We leave more complicated selection methods for future work.

\begin{figure}[t]
    \centering
    \includegraphics[width=0.7\linewidth]{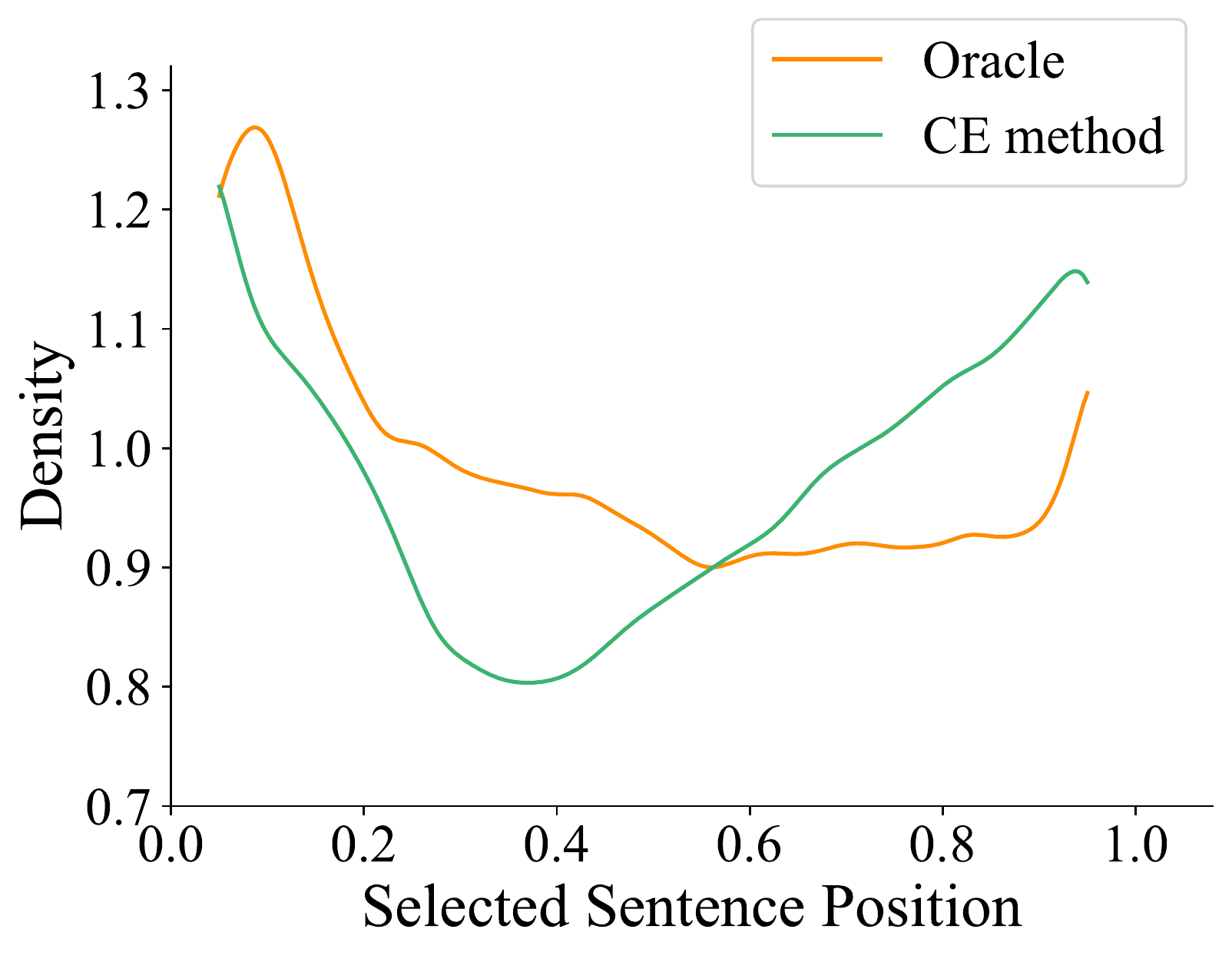}

    \caption{Selected sentence position distribution. We use the relative position of each sentence with regard to the whole article, thus taking values from 0 to 1.
    }
    \label{fig: selected_dist}
\end{figure}

\paragraph{Oracle Extraction}
First, for comparison purposes, we construct an oracle for each paper which is the extraction that achieves highest average ROUGE scores with respect to reference reviews, specifically using the greedy method described in \citet{Nallapati2017SummaRuNNerAR}. Note that for each paper with multiple reviews, we construct multiple oracles for that paper. We assume that oracle extractions can reflect where reviewers pay more attention to when they are writing reviews.
The selected sentence position distribution in oracles is shown in Fig.~\ref{fig: selected_dist}.

\paragraph{Section-based Extraction}
Scientific papers are highly structured. As a convention, a scientific paper usually describes problem background, related work comparison, as well as its own contributions in the introduction part.
Regarding this method, we only use the introduction section, which can be regarded as a baseline model.

\paragraph{Cross-entropy (CE) Method Extraction}
\label{para:ce}
Here we select salient sentences from the full text range. The way we do so is through a two-step selection process:
\begin{enumerate*}
    \item Select sentences containing certain informative keywords (e.g. propose) which are detailed in Appendix~\ref{appendix:cem}. Those selected sentences form a set $\mathcal{S}$.
    \item Select a subset $\mathcal{S'} \subseteq \mathcal{S}$ such that sentences in $\mathcal{S'}$ cover diverse content and satisfy a length constraint.
\end{enumerate*}
In the second step, we use the cross-entropy method introduced in \citet{10.1145/3077136.3080690} where we select diverse content by maximizing unigram entropy. The details of this two-step process can be found in Appendix \ref{appendix:cem}.
The selected sentence position distribution using this method is shown in Fig.~\ref{fig: selected_dist}.
We can see that the extractor tends to select sentences from the beginning of a paper as well as the ending part of a paper just as the oracle extractor does. This makes sense because the beginning part is the introduction part which talks about the essence of the whole paper and the ending part mostly contains the analysis of experimental results and conclusions etc.

\paragraph{Hybrid Extraction}
We combine the abstract of a paper and its CE extraction to form a hybrid of both.

\subsubsection{Aspect-aware Summarization}

\begin{figure}[h]
    \centering
    \includegraphics[width=0.9\linewidth]{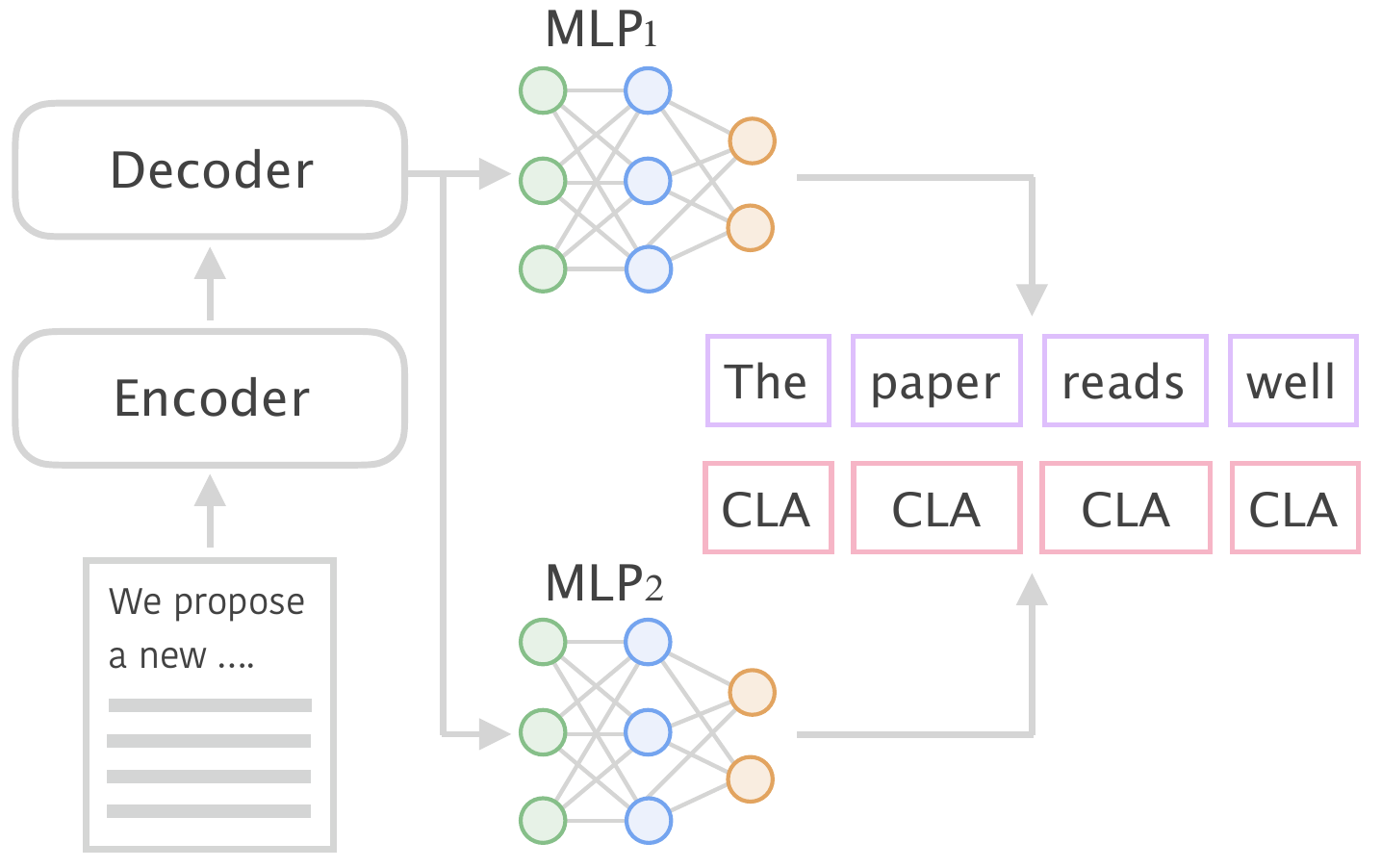}
    \caption{Aspect-aware summarization.
    }
    \label{fig: arch}
\end{figure}

Typically in the \textit{extract-then-generate} paradigm, we can just use the extractions directly and build a sequence-to-sequence model to generate text.
Here, in order to generate reviews with more diverse aspects and to make it possible to interpret the generated reviews through the lens of their internal structure, we make a step towards a generation framework involving \textit{extract-then-generate-and-predict}.
Specifically, instead of existing aspect-based summarization works that explicitly take aspects as input \cite{angelidis-lapata-2018-summarizing,frermann-klementiev-2019-inducing,hayashi20tacl}, we use our annotated aspects (\S\ref{sec:aspect-enhanced}) as additional information, and design an auxiliary task that aims to predict aspects of generated texts (reviews).
Fig.~\ref{fig: arch} illustrates the general idea of this.

The loss of this model is shown in Eq.~\ref{eq:1}
\begin{equation} \label{eq:1}
    \mathcal{L} = \mathcal{L}_{\text{seq2seq}} + \alpha\mathcal{L}_{\text{seqlab}}
\end{equation}
where $\mathcal{L}_{\text{seq2seq}}$ denotes sequence to sequence loss which is the negative log likelihood of the correct next tokens, and $\mathcal{L}_{\text{seqlab}}$ denotes sequence labeling loss which is the negative log likelihood of the correct labels of next tokens. $\alpha$ is a hyper-parameter ($\alpha = 0.1$) that is tuned to maximize aspect coverage on the development set.

\section{Experiment}

In this section, we investigate using our proposed review generation systems with state-of-the-art pre-trained models, to what extent can we realize desiderata of reviews that we defined in \S\ref{sec: multi-pers-eval}.
We approach this goal by two concrete questions:
(1) \textit{What are review generation systems (not) good at?}
(2) \textit{Will systems generate biased reviews?}

\subsection{Settings}

Here we consider three extraction strategies in \S\ref{sec:2stage} as well as two generation frameworks, one is the vanilla sequence to sequence model, the other is jointly sequence to sequence and sequence labeling.

\paragraph{Dataset}
We use our constructed dataset \texttt{ASAP-Review} described in \S\ref{sec: dataset} to conduct experiments.
For each paper, we use full text (without Appendix) as source document.\footnote{If a paper has more than 250 sentences, we truncate it and take the first 250 sentences when we do the extraction step.} And we filtered papers with full text fewer than 100 words since they don't contain enough information for models to learn.
For reviews, we only use 100-1024 word reviews \footnote{As measured by BART's subword tokenizer.} for training due to computational efficiency, which account for $92.57\%$ of all the reviews.
This results in 8,742 unique papers and 25,986
paper-review pairs in total, the split of our dataset is shown in Tab.~\ref{tab:split}.

\begin{table}[h]
\small
    \centering
    \begin{tabular*}{0.43\textwidth}{@{}lcccc@{}}
\toprule
\hspace{3mm} & Train & Validation & Test  \\ \midrule
\hspace{3mm}Unique papers & 6,993 & 874 & 875 \\
\hspace{3mm}Paper-review pairs & 20,757  & 2,571  & 2,658 \\  \bottomrule
\end{tabular*}
    \caption{Data split of \texttt{ASAP-Review}.}
    \label{tab:split}
\end{table}

\paragraph{Model}

\begin{table*}[h]
\small
\renewcommand{\arraystretch}{1.15}
\begin{center}
\setlength\tabcolsep{5pt}
\begin{tabular*}{0.98\textwidth}{l c c c c c c c c c c c}
\toprule
\textbf{Desiderata} & & \textbf{Decisive.} & \multicolumn{2}{c}{\textbf{Comprehen.}} & \textbf{Justification} & \multicolumn{2}{c}{\textbf{Accuracy}} & \multicolumn{4}{c}{\textbf{Others}}\\ \midrule

\textbf{Metric} & & \textbf{\textsc{Racc}} & \textbf{\textsc{Acov}} & \textbf{\textsc{Arec}} & \textbf{\textsc{Info}} & \textbf{\textsc{Acon}} & \textbf{\textsc{Sacc}} & \textbf{R-1} & \textbf{R-2} & \textbf{R-L} & \textbf{BS}  \\ \midrule
HUMAN & & 30.32 & 49.85  & 58.66  & 97.97 & 75.67 & 90.77 & -- & -- & -- & --  \\ \midrule
\multicolumn{11}{c}{\textsc{Extractive}} \\ \midrule

INTRO  & & -- & -- & -- & -- & -- & -- & 38.62 &  8.84 & 25.11 & 29.22 \\
CE & & -- & -- & -- & -- & -- & -- & 38.56 & 7.81 & 25.94 & 29.11 \\
ABSCE & & -- & -- & -- & -- & -- & -- & 37.55 & 8.53 & 25.85 & 31.99\\ \midrule
\multicolumn{11}{c}{\textsc{Extractive+Abstractive}}\\
\midrule
& Aspect & & & \\ \midrule

\multirow{2}{*}{INTRO}  & $\times$  & -15.38$\dag$ & 50.37  & 55.52$\dag$ & 100.00$\dag$ & 43.78$\dag$ & 83.93 & 41.39 & 11.53 & 38.52 & 42.29 \\
\multirow{2}{*}{} & $\surd$ & -11.54$\dag$ & 51.50 & 58.24 & 99.29 & 32.51$\dag$ & 80.36$\dag$ & 41.31 & 11.41 & 38.38 & 42.33 \\ \midrule

\multirow{2}{*}{CE}  & $\times$ & -23.08$\dag$ & 62.64$\dag$  & 60.73 & 99.29 & 39.17$\dag$ & 78.57$\dag$ & 42.37 & 11.72 & 39.86 & 41.78  \\
\multirow{2}{*}{} & $\surd$ & -30.77$\dag$ & 63.96$\dag$ & 61.62$\dag$ & 100.00$\dag$ & 34.46$\dag$ & 69.64$\dag$ & 42.27 & 11.62 & 39.73 & 41.71  \\ \midrule

\multirow{2}{*}{ABSCE}  & $\times$ & -30.77$\dag$ & 55.37$\dag$  & 58.31 & 98.21 & 34.75$\dag$ & 92.86 & 43.11 & 12.24 & 40.18 & 42.90  \\
\multirow{2}{*}{} & $\surd$ & -38.46$\dag$ & 56.91$\dag$ & 57.56 & 98.21 & 35.21$\dag$ & 87.50 & 42.99& 12.19& 40.12 & 42.63 \\
\bottomrule
\end{tabular*}
\end{center}

\caption{\label{performance-table} Results of the baseline models as well as different aspect-enhanced models under diverse automated evaluation metrics. ``BS'' represents BERTScore. $\dag$ denotes that the difference between system generated reviews and human reviews are statistically significant (p-value  $<$ 0.05 using 10,000 paired bootstrap resampling \cite{efron1992bootstrap} tests with 0.8 sample ratio).}
\end{table*}

As mentioned in \S\ref{sec:sysdesign}, the pre-trained sequence-to-sequence model we used is BART.
For all models, we initialized the model weights using the checkpoint: ``\texttt{bart-large-cnn}'' which is pre-trained on ``\texttt{CNN/DM}" dataset \citep{hermann2015teaching}.\footnote{We also tried ``\texttt{bart-large-xsum}" checkpoint which is pre-trained on ``\texttt{XSUM} dataset \citep{xsum-emnlp}", however that results in much shorter reviews, and sentences in it tend to be succinct.} For \textit{extract-then-generate-and-predict} framework, we add another multilayer perceptron on top of the BART decoder, and initialize it with 0.0 mean and 0.02 standard deviation. We use the Adam optimizer\citep{kingma2014adam} with a linear learning rate scheduler which increases the learning rate linearly from 0 to $4e^{-5}$ in the first $10\%$ steps (the warmup period) and then decreases the learning rate linearly to 0 throughout the rest of training steps. We finetuned our models on the whole dataset for 5 epochs. We set a checkpoint at the end of every epoch and finally took the one with the lowest validation loss.

During generation, we used beam search decoding with beam size 4. Similarly to training time, we set a minimum length of 100 and a maximum length of 1024. A length penalty of 2.0 and trigram blocking \citep{paulus2017deep} were used as well.

\subsection{What are Systems Good and Bad at?}
\label{sec:results}
Based on the evaluation metrics we defined in \S\ref{sec: multi-pers-eval}, we conduct both automatic evaluation and human evaluation to characterize both reference reviews and generated reviews, aiming to analyze what sub-tasks of review generation automatic systems can do passably at, and also where they fail. The aspect information in each review is obtained using aspect tagger we trained in \S\ref{sec:aspect-enhanced}.

\paragraph{Automatic Evaluation}
Automatic evaluation metrics include \textit{Aspect Coverage} (\textsc{ACov}), \textit{Aspect Recall} (\textsc{ARec}) and \textit{Semantic Equivalence} (ROUGE, BERTScore). Notably, for each source input, there are multiple reference reviews. When aggregating ROUGE and BERTScore\footnote{We have used our own custom baseline to rescale BERTScore, details can be found in Appendix \ref{app:detail-metric}.}, we take the maximum instead of average. And when aggregating other metrics for human reviews, we take the average for each source document. The results are shown in Tab.~\ref{performance-table}.

\paragraph{Human Evaluation}
Metrics that require human labor include \textit{Recommendation Accuracy} (\textsc{RAcc}), \textit{Informativeness} (\textsc{Info}), \textit{Aspect-level Constructiveness} (\textsc{ACon}) and \textit{Summary Accuracy} (\textsc{SAcc}). We select 28 papers from ML/NLP/CV/RL domains. None of these papers are in the training set. Details regarding human judgment are shown in Appendix \ref{app:detail-metric}. The evaluation results are shown in Tab.~\ref{performance-table}.

Overall, we make the following observations:

\subsubsection{Weaknesses}

Review generation system will generate non-factual statements for many aspects of the paper assessment, which is a serious flaw in a high-stakes setting.

\paragraph{Lacking High-level Understanding}
Specifically, when using metrics that require higher level understanding of the source paper like \textit{Recommendation Accuracy} and \textit{Aspect-level Constructiveness}, proposed systems achieved much lower performance, with even the smallest gaps between systems and humans being $41.86\%$ for \textit{Recommendation Accuracy} and $31.89\%$ for \textit{Aspect-level Constructiveness} compared to reference reviews. This means our systems cannot precisely distinguish high-quality papers from low-quality papers and the evidence for negative aspects is not reliable most of the time.\footnote{Although there exist varying degrees of performance differences on \textbf{RACC} and \textbf{ACon} for different systems, we only find one pair of systems perform statistically different on \textbf{ACon}.}

\paragraph{Imitating Style}
After careful inspection, we find that some of sentences will appear frequently in different generated results.
For example,  the sentence ``\texttt{The paper is well-written and easy to follow}" appears in more than $90\%$ of generated reviews due to the fact that in the training data, this exact sentence appears in more than $10\%$ of papers. This suggests that the style of generated reviews tend to be influenced by high-frequency sentence patterns in training samples.

\paragraph{Lack of Questioning}
Generated reviews ask few questions about the paper content, which is an important component in peer reviewing. In the reference reviews, the average number of questions per review is 2.04, while it is only 0.32 in generated reviews.

\subsubsection{Advantages}
We find that review generation systems can often precisely summarize the core idea of the input paper, and generate reviews that cover more aspects of the paper's quality than those created by human reviewers. Systems with aspect information are also aspect-aware and evidence sensitive as we will discuss below.

\paragraph{Comprehensiveness}
In terms of \textit{Aspect Coverage} and \textit{Informativeness}, our systems can outperform human reviewers by at most $14.11\%$ and $2.03\%$ respectively, suggesting that even reviews from the reviewers may also fall short on our defined criterion regarding comprehensiveness.

\paragraph{Good Summarization}
Current systems can correctly summarize the contributions of papers most of the time as shown by \textit{Summary Accuracy}. $4$ out of $6$ systems can achieve over $80\%$ accuracy and statistical significance tests show that gaps between top-$3$ systems and human reviewers are not significant. This means that in terms of summarizing the paper content, current systems can achieve comparable performance to human reviewers.

\subsubsection{System Comparisons}

We also look into how systems with different settings are diverse in performance and make the following observations.

\paragraph{Summarization Paradigms}
By looking at ROUGE (R) and BERTScore (BS), we can see that ``extractive + abstractive''-based methods can consistently outperform pure extractive methods, with the smallest gaps of $2.69$, $2.57$, $12.44$, $9.72$ for R-1, R-1, R-L and BS respectively. This demonstrates the necessity of using abstractive summarization which can generate reviews that are close both in meaning as well as language use to human reviews.

\paragraph{Extraction Strategies}
We can see that it is more effective to use extracted text from the full paper to aid the generation process, resulting in higher aspect coverage compared with solely using introduction information.
This is reasonable since models can obtain more diverse input from the full text.

\begin{table*}[!htbp]
\small
\begin{center}
\renewcommand{\arraystretch}{1.5}
\begin{tabular}{p{0.95\textwidth}}
\toprule
\mycbox{sum} summary \qquad\qquad\quad
\mycbox{cla} clarity + \qquad\qquad\quad
\mycbox{sub} substance + \qquad\qquad\quad
\mycbox{sou} soundness - \qquad\qquad\quad
\mycbox{ori} substance -
\\ \midrule
\setlength{\fboxsep}{0pt}\colorbox{sum}{This paper studies the problem of transfer learning from a single pre-trained network onto a new task . The authors pro-}\\
\setlength{\fboxsep}{0pt}\colorbox{sum}{pose a method of combining the outputs of multiple pre-trainable classifiers by training on their raw predictions and then}\\
\setlength{\fboxsep}{0pt}\colorbox{sum}{fine-tuning on the target samples . The proposed method is based on the use of ( non-linear ) maximal correlation analy-}\\
\setlength{\fboxsep}{0pt}\colorbox{sum}{sis that originated with Hirschfeld [ 9 ] to this problem . The paper is well}\setlength{\fboxsep}{0pt}\colorbox{cla} {written and easy to follow . The} experimental\\
results \setlength{\fboxsep}{0pt}\colorbox{sub}{show that} the proposed method outperforms the state-of-the-art methods on the CIFAR-100 , Stanford Dogs , and\\
Tiny ImageNet datasets . However , it is not clear to me how the performance of the method is affected \setlength{\fboxsep}{0pt}\colorbox{sou}{by} the number of \\
target training samples . It would be better if \setlength{\fboxsep}{0pt}\colorbox{ori}{the authors} can provide \setlength{\fboxsep}{0pt}\colorbox{ori}{some theoretical analysis on the} effect of the size of\\ \setlength{\fboxsep}{0pt}\colorbox{ori}{the} target dataset .
\\
\bottomrule
\end{tabular}
\end{center}

\caption{\label{s2sal} Illustration of generated tokens associated with corresponding aspects. + denotes positive sentiment. - denotes negative sentiment.}

\end{table*}

\paragraph{System Diagnosis}
Our fine-grained evaluation metrics enable us to compare different systems and interpret their relative merits. For example, as discussed before, our systems can achieve higher \textit{Informativeness} than reference reviews while suffering from much lower \textit{Aspect-level Constructiveness}. This means that if we want our systems to match the performance of real reviewers, we should focus on improving the constructiveness of our systems instead of aiming for methods that provide better evidence for negative aspects (which are not factually correct most of the time in the first place).

\subsubsection{Case Study}

To get an intuitive understanding of how aspect-enhanced review generation system worked, we perform analysis on a real case. (More analysis can be found in Appendix \ref{app:detail_result_analysis}.)
Specifically, since our aspect-enhanced model is trained based on multi-task learning framework, it would be interesting to see how well the tokens are generated associated with corresponding aspects.
We take our aspect-enhanced model using CE extraction to conduct this experiment. Tab.~\ref{s2sal} shows an example review when we do so.

We can see that the model can not only generate fluent text but also be aware of what aspect it is going to generate as well as the correct polarity of that aspect. Although the generated aspects are often small segments and there are some minor alignment issues, the model is clearly aspect-aware.

\subsection{Will System Generate Biased Reviews?}
\label{sec:bias-analysis}

Biases in text are prevalent, but often challenging to detect \cite{manzoor2020uncovering,stelmakh2019testing}. For example, in natural language processing, researchers are trying to identify societal biases (e.g, gender) in data and learning systems on different tasks \cite{bolukbasi2016man,zhao2018gender,stanovsky-etal-2019-evaluating}.
However, previous works on analyzing bias in scientific peer review usually focus on disparities in numerical feedback instead of text.
\citet{manzoor2020uncovering} recently uncover latent bias in peer review
based on review text. In this work,  besides designing a model to generate reviews, we  also perform an analysis of bias, in which we propose a method to identify and quantify biases both in human-labeled and system-generated data in a more fine-grained fashion.

\subsubsection{Measuring Bias in Reviews}
\label{sec:measure_bias}

\begin{figure}[t]
    \centering
    \includegraphics[width=1\linewidth]{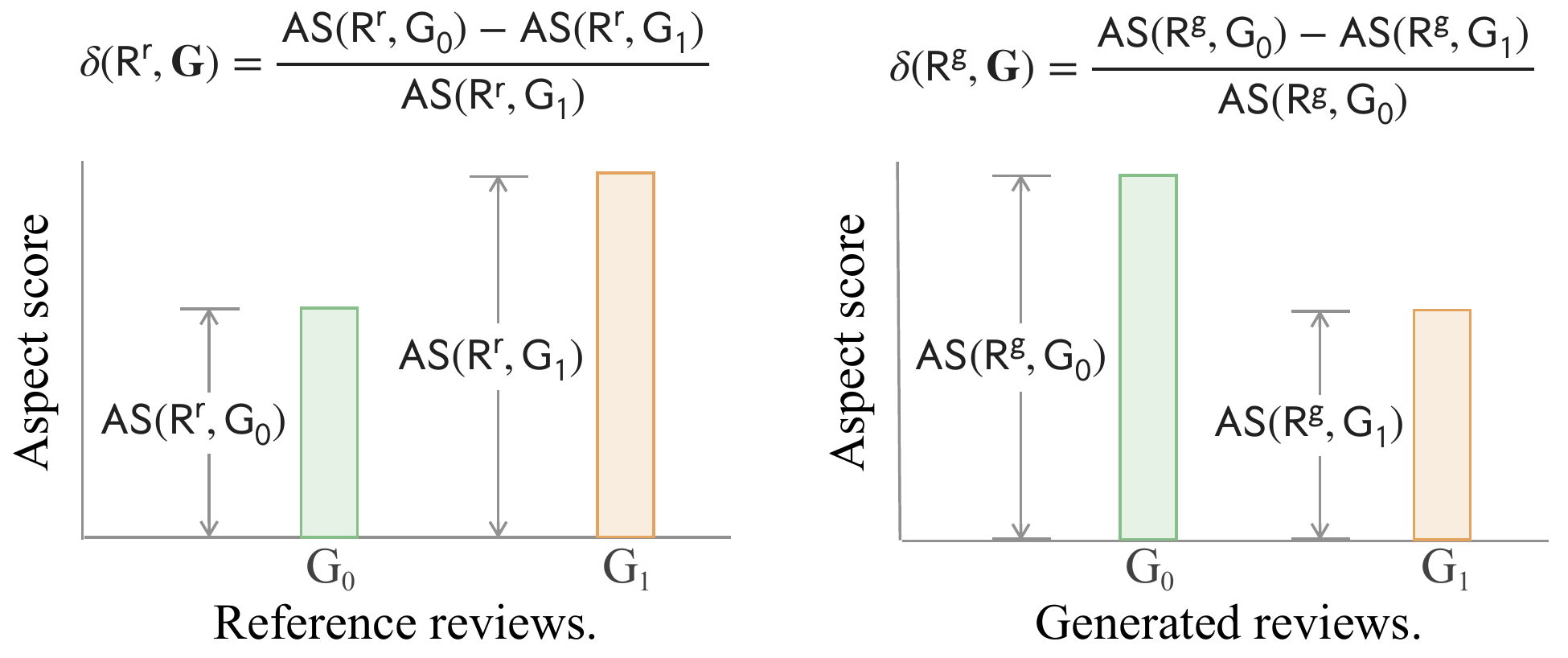}
    \caption{\textit{Aspect score} $\textsc{AS}(R, G_i)$ and \textit{disparity} $\delta(R,\mathbf{G})$ in reference reviews ($R^r$) and generated reviews ($R^g$). $\mathbf{G} = [G_0, G_1]$ denotes different groups.}
    \label{fig:bias-vis}
\end{figure}

To characterize potential biases existing in reviews, we (i) first define an \textit{aspect score}, which calculates the percentage of positive occurrences\footnote{If an aspect does not appear in a review, then we count the score for that aspect 0.5 (stands for neutral). Details see Appendix \ref{app:aspect_score}.} of each aspect. The polarity of each aspect is obtained based on our learned tagger in \S\ref{sec:aspect-enhanced};
(ii) then we aim to observe if different groups $G_i$ (e.g., groups whether the paper is anonymous during reviewing or is not anonymous) of reviews $R$ would exhibit \textit{disparity} $\delta(R,\mathbf{G})$ in different aspects. The calculation of disparity can be visualized in Fig.~\ref{fig:bias-vis}.

Based on above two definitions, we characterize bias in two ways respectively:

\noindent (1) \textbf{spider chart}, which directly visualizes aspect scores of different groups of reviews w.r.t~each aspect.

\noindent (2) \textbf{disparity difference}, which represents the difference between disparities in generated reviews $R^g$ and reference reviews $R^r$ and can be formally calculated as:
\begin{align}
        \Delta(R^g, R^r, \mathbf{G}) = \delta(R^g, \mathbf{G}) - \delta(R^r, \mathbf{G})
\end{align}
where $\mathbf{G} = [G_0, G_1]$ denotes different groups based on a given partition criterion. Positive value means generated reviews favor group $G_0$ more compared to reference reviews, and vice versa.

In this work, we group reviews from two perspectives. The basic statistics are shown in Tab.~\ref{tab:bias-stat}.

\begin{table}[h]
\footnotesize
    \centering
    \setlength\tabcolsep{4pt}
    \begin{tabular*}{0.48\textwidth}{@{}lcccc@{}}
\toprule
\hspace{2mm}                            & Native & Non-native & Anonym. & Non-anonym. \\ \midrule
\hspace{2mm}Total           & 651  & 224 & 613 & 217 \\
\hspace{2mm}Acc.\%     & 66.51\% & 50.00\% & 57.59\% & 78.34\%  \\
\bottomrule
\end{tabular*}
    \caption{Test set statistics based on nativeness and anonymity.}
    \label{tab:bias-stat}
\end{table}

\begin{figure*}[!htbp]
    \centering
    \includegraphics[width=0.95\linewidth]{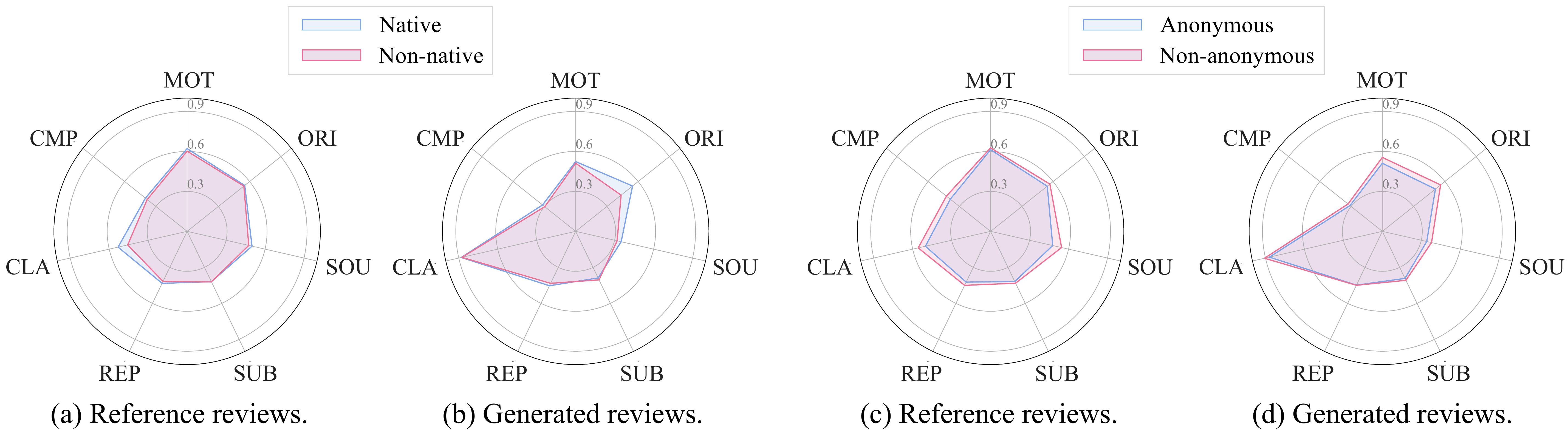}
    \caption{Spider chart of aspect scores with respect to different groups.}
    \label{fig:all-bias}
\end{figure*}

\begin{table*}[htp]
\centering
\setlength\tabcolsep{12pt}

\small
\renewcommand{\arraystretch}{1.2}
\begin{tabular*}{0.95\textwidth}{@{}lcccccccc@{}}

\toprule
\hspace{4mm}  & MOT & ORI & SOU & SUB & REP & CLA & CMP & Total\\ \midrule
\hspace{4mm}Nativeness & -0.72 & +18.71 & +3.84 & -3.66 & +0.73 & -13.32 & +2.40 & 43.39
\\
\hspace{4mm}Anonymity & -5.69 & -4.43 & +2.76 & -0.64 & +5.65 & +5.80 & +3.02 & 28.00 \\

\bottomrule
\end{tabular*}
\caption{\label{aspect-bias-intro} Disparity differences regarding nativeness and anonymity. Total is the sum of absolute value of disparity difference.}

\end{table*}

\paragraph{Nativeness}
We categorize all papers in test set into ``native" ($G_0$) and ``non-native" ($G_1$) based on whether there is at least one native speaker in the author list as well as whether the institution is in an English-speaking country.\footnote{We used \url{https://www.familysearch.org/en/} to decide the nationality of an author. In cases where all authors are not from an english-speaking country, we look into the institution information to further decide the attribution of the paper based on whether the institution is from an english-speaking country.}

\paragraph{Anonymity}
We categorize all papers in test set into ``anonymous" ($G_0$) and ``non-anonymous" ($G_1$) based on whether the paper has been released as a pre-print before a half month after the conference submission deadline.\footnote{We discard papers from ICLR 2017 since the reviewing process was single blind.}

Here we take our model with introduction extraction as an example to showcase how to use the fine-grained aspect information in our dataset to do bias analysis. We list the bias analysis for other models in Appendix~\ref{app: bias-analysis}.

\subsubsection{Nativeness Analysis}
\paragraph{Spider Chart}
Generally, \texttt{Native} papers receive higher score in most aspects in both reference reviews and generated reviews. Specifically, for human reviews:
(1) By looking at Fig.~\ref{fig:all-bias}-(a), there is a significant gap in \texttt{Clarity}, which is reasonable since non-native authors may have more trouble conveying their ideas.
(2) Scores of the two groups are much closer in other aspects.

For system-generated reviews:
As shown in Fig.~\ref{fig:all-bias}-(b), the auto-review system narrows the disparity in \texttt{Clarity} but amplifies it in \texttt{Originality}, meaning that system reviewers are harsher than human reviewers when commenting the paper’s    ``\texttt{Originality}'' for non-native English speakers.
This observation suggests that a review system can generate biased reviews in some aspects, which would lead to unfair comments. Therefore, a system should be de-biased before it come to use.

\paragraph{Disparity Difference}
Through spider chart, gaps between different groups are relatively small and hard to discern. Besides, those gaps can only show the absolute favor for a certain group in different aspects. We are also interested in whether generated reviews are more in favor of a certain group \textbf{compared to reference reviews}. To do this, we calculate disparity differences and list them in Tab.~\ref{aspect-bias-intro}.

As shown in Tab.~\ref{aspect-bias-intro}, for \texttt{Originality} and \texttt{Clarity}, the disparity difference is $+18.71$ and $-13.32$ which means that the system favours native papers in \texttt{Originality} and non-native papers in \texttt{Clarity} \textbf{compared to} human reviewers. This observation is consistent with spider chart. Besides, varying degrees of bias are presented in Tab.~\ref{aspect-bias-intro}. For example, for \texttt{Motivation} and \texttt{Replicability}, the disparity difference is less than $1$, which suggests little bias while in other aspects, the bias is much larger.

\subsubsection{Anonymity Analysis}
\paragraph{Spider Chart}
By looking at Fig.~\ref{fig:all-bias}-(c) and Fig.~\ref{fig:all-bias}-(d), we find that both human reviewers and system reviewers favor non-anonymous papers in all aspects.
Specifically, for human reviews: we find gaps are non-negligible in \texttt{Soundness}, \texttt{Clarity} and \texttt{Meaningful Comparison} while for system-generated reviews,
we observe that gaps are considerable in \texttt{Motivation}, \texttt{Originality}, \texttt{Soundness}.
This observation is interesting since human reviewers may be aware of the identity of the authors due to non-anonimity which may affect the reviews they write. However, our system is not aware of that and its preference towards non-anonymous paper probably suggests some quality difference.\footnote{Non-anonymous papers are more likely to have been rejected before and therefore are revised many more times.}

\paragraph{Disparity Difference}
By looking at Tab.~\ref{aspect-bias-intro}, we find that the largest absolute disparity difference regarding anonymity is $5.80$ compared to $18.71$ regarding nativeness. This suggests that regarding anonymity, our system's preference does not diverge that much from human reviewers. Also, the total aspect bias regarding anonymity is $28.00$, much smaller compared to total aspect bias regarding nativeness ($43.00$). This also suggests that our model is less sensitive to anonymity compared to nativeness.

The observations above are probably related to some superficial heuristics existing in peer review. For example, when reviewers detect some grammar mistakes, they may assume that the authors are not native and then bias towards rejecting the paper by claiming some clarity issues. Another example is that there may exist differences in the research topics pursued by different subgroups (e.g., different countries), the bias regarding nativeness may also suggest the favor of certain topics in the reviewing process. Those superficial heuristics should be discouraged and deserve further investigation in future research.

\section{Related Work}
\paragraph{Scientific Review Generation}
There has been a relative paucity of work on scientific review generation, other than \citet{bartoli2016your}'s work investigating the feasibility of generating fake reviews by surface-level term replacement and sentence reordering etc.
In addition contemporaneous and independent work by \citet{wang2020reviewrobot} proposes a two-stage information extraction and summarization pipeline to generate paper reviews.
Their evaluation focuses mainly on the accuracy of information extraction, and the evaluation of the generated summaries is somewhat precursory, assessing only a single criterion ``constructiveness and validity'' manually over 50 papers.
Our paper (1) proposes a wide variety of diagnostic criteria on review quality, (2) uses a very different summarization methodology, and (3) evaluates the generated results extensively.

\paragraph{Peer Review}
Peer review is an essential component of the research cycle and is adopted by most journals and conferences to identify important and relevant research. However, at the same time it is easy to identify many issues: expensiveness, slowness, existence of inconsistency \citep{langford2015arbitrariness} and bias \citep{tomkins2017reviewer}, etc.

Some efforts have been put into analyzing the peer review process including automating review assignment \citep{jin2017integrating, nguyen2018decision, anjum-etal-2019-pare,jecmen2020mitigating}, examining bias problems \citep{tomkins2017reviewer, stelmakh2019testing}, examining consistency problems \citep{langford2015arbitrariness} and performing sentiment analysis on reviews \citep{wang2018sentiment,chakraborty2020aspect}.
Several decision classification methods have been explored to help make accept or reject decision given a paper. Those methods are either based on textual \citep{kang18naacl, qiao2018modularized} or visual \citep{von2010paper, huang2018deep} information.
However, they do not directly alleviate review load, as our paper aims to do.

\section{Discussion and Future Directions}
We first summarize what we have achieved in this work and how the current \textit{ReviewAdvisor} system can potentially help in a reviewing process. Then we discuss challenges and potential directions for the automatic review generation task, which, hopefully,  encourages more future researchers to explore this task, and in the right direction.

\subsection{Machine-assisted Review Systems} \label{sec:assist}
Instead of replacing a human reviewer, a better position for \textit{ReviewAdvisor} is to regard it as a machine-assisted review system.
Although there is still a large room for improvement, our results indicate that even with current technology:

(1) Based on the evaluation of \S\ref{sec:results}, \textit{Summary Accuracy} of our systems is quite high, suggesting that it can be either used for reviewers to finish the description of \textit{Summary}, or help general readers to quickly understand the core idea of recently pre-printed papers (e.g., papers from arXiv).

(2) Based on evaluation of \S\ref{sec:results}, reviews generated by \textit{ReviewAdvisor} can cover more aspects and generate more informative reviews. Although the associated opinions may suffer from  constructiveness problems, they still may be useful since they can provide a preliminary template for reviewers, especially enabling junior or non-native English reviewers to know what a review generally should include and how to phrase each aspect.
Additionally, for each aspect (e.g., \textit{Clarity}), our system can provide relevant evidence sentences from the paper, helping reviewers quickly identify salient information when reviewing the paper (Detailed example in our Appendix \ref{app:detail_result_analysis}).

\subsection{Challenges and Promising Directions} \label{sec:challenges}
\subsubsection{Model}

 \noindent (1) \textit{Long Document Modeling}:
The average length of one scientific paper is commonly larger than 5,000 words, far beyond the input text's length that mainstream neural sequence models (e.g., LSTM, Transformer) or pre-trained models (e.g., BERT, BART) normally use. This work (in \S\ref{sec:2stage}) bypasses the difficulty by using a two-stage system, but other strategies should be explored.

 \noindent (2) \textit{Pre-trained Models for Scientific Domain}:
Although previous works, as exemplified by \cite{beltagy2019scibert} have pre-trained BERT on scientific domain, we observe that using these models with transformer decoders perform much worse than BART on sequence generation tasks in terms of fluency and coherence, which calls for general sequence to sequence models pre-trained on scientific domain for higher-quality review generation.

\noindent (3) \textit{Structure Information}:
Review generation systems could get a deeper understanding of a given research paper if structural information can be provided.
To this end, outputs from scientific paper-based information extraction tasks \cite{hou-etal-2019-identification,jain-etal-2020-scirex} can be utilized to guide review generation.

\noindent (4) \textit{External Knowledge}:
Besides the paper itself, review systems can also rely on external knowledge, such as a citation graphs constructed based on more scientific papers or a knowledge graph connecting concepts across different papers \citep{luan2018multi, lo-wang-2020-s2orc}.
Also,  recently, \citet{august-etal-2020-writing} compile a set of writing strategies drawn from a
wide range of prescriptive sources, it would be also valuable to transfer this knowledge into the auto-review system.

\subsubsection{Datasets}
 \noindent (5) \textit{More Open, Fine-grained Review Data}: In this work, we annotate fine-grained information (aspects) of each review manually. However, this information could potentially be obtained directly from the peer review system. How to access this information appropriately would be an important and valuable step in the future.

 \noindent (6) \textit{More Accurate and Powerful Scientific Paper Parsers}:
 Existing parsing tools (e.g. science-parse, grobid) for scientific papers are commonly designed for certain specific paper templates, and also still struggle at extracting fine-grained information, such as the content of tables and figures.

\subsubsection{Evaluation}
 \noindent (7) \textit{Fairness and Bias in Generated Text}:
 In this work, we make a step towards identifying and quantifying two types of biases existing in human and system-generated reviews. Future works can explore more along this direction based on our dataset that contains  fine-grained aspect annotation.

 \noindent (8) \textit{Factuality and Reliability}:
 A generated review should be \textit{factually correct} \cite{wadden-etal-2020-fact} which also poses challenge to the current evaluation methodology.
In addition to generating a review, a reliable system should also provide a level of confidence with respect to the current comment.
Moreover, whether review scores are calibrated is another valuable question.

\subsection{Conclusion}

In answer to the titular question of ``can we automate scientific review,'' the answer is clearly ``not yet''.
However, we believe the models, data, and analysis tools presented in this paper will be useful as a starting point for systems that can work in concert with human reviewers to make their job easier and more effective.

\section*{Acknowledgment}

This work could not be accomplished without the help of many researchers.
We would like to thank people for their generous support, especially,

\paragraph{Volunteer to help us with the human evaluation:}
Gábor Berend, Zhouhan Lin, William W. Cohen, Pengcheng Yin, Tiange Luo, Yuki M. Asano, Junjie Yan, Tuomas Haarnoja, Dandan Guo, Jie Fu, Lei Chen, Jinlan Fu, Jiapeng Wu, Wenshan Wang, Ziyi Dou, Yixin Liu, Junxian He, Bahetiyaer Bare, Saizheng Zhang, Jiateng Xie, Spyros Gidaris, Marco Federici, Junji Dai, Zihuiwen Ye Jie Zhou, Yufang Liu, Yue Zhang, Ruifeng Xu, Zhenghua Li, Chunting Zhou, Yang Wei.

This work lasted nearly a year, from the initial idea discussion (2020.02.28) to completing the first version of draft (2021.01.29).
This is the year from the beginning of the COVID-19 epidemic to its outbreak. Thanks for this fun and challenging project that punctuates my dull life at home. Thank Weizhe, for her patience, persistence and her willingness to work with me to complete this crazy idea.
Thanks a lot for Graham's endless help on this project.
The story is not over, and our system is still evolving.

\bibliography{emnlp2020}

\begin{thebibliography}{74}
\expandafter\ifx\csname natexlab\endcsname\relax\def\natexlab#1{#1}\fi

\bibitem[{Angelidis and Lapata(2018)}]{angelidis-lapata-2018-summarizing}
Stefanos Angelidis and Mirella Lapata. 2018.
\newblock \href {https://doi.org/10.18653/v1/D18-1403} {Summarizing opinions:
  Aspect extraction meets sentiment prediction and they are both weakly
  supervised}.
\newblock In \emph{Proceedings of the 2018 Conference on Empirical Methods in
  Natural Language Processing}, pages 3675--3686, Brussels, Belgium.
  Association for Computational Linguistics.

\bibitem[{Anjum et~al.(2019)Anjum, Gong, Bhat, Hwu, and
  Xiong}]{anjum-etal-2019-pare}
Omer Anjum, Hongyu Gong, Suma Bhat, Wen-Mei Hwu, and JinJun Xiong. 2019.
\newblock \href {https://doi.org/10.18653/v1/D19-1049} {{P}a{R}e: A
  paper-reviewer matching approach using a common topic space}.
\newblock In \emph{Proceedings of the 2019 Conference on Empirical Methods in
  Natural Language Processing and the 9th International Joint Conference on
  Natural Language Processing (EMNLP-IJCNLP)}, pages 518--528, Hong Kong,
  China. Association for Computational Linguistics.

\bibitem[{August et~al.(2020)August, Kim, Reinecke, and
  Smith}]{august-etal-2020-writing}
Tal August, Lauren Kim, Katharina Reinecke, and Noah~A. Smith. 2020.
\newblock \href {https://doi.org/10.18653/v1/2020.emnlp-main.429} {Writing
  strategies for science communication: Data and computational analysis}.
\newblock In \emph{Proceedings of the 2020 Conference on Empirical Methods in
  Natural Language Processing (EMNLP)}, pages 5327--5344, Online. Association
  for Computational Linguistics.

\bibitem[{Bartoli et~al.(2016)Bartoli, De~Lorenzo, Medvet, and
  Tarlao}]{bartoli2016your}
Alberto Bartoli, Andrea De~Lorenzo, Eric Medvet, and Fabiano Tarlao. 2016.
\newblock Your paper has been accepted, rejected, or whatever: Automatic
  generation of scientific paper reviews.
\newblock In \emph{International Conference on Availability, Reliability, and
  Security}, pages 19--28. Springer.

\bibitem[{Beltagy et~al.(2019)Beltagy, Lo, and Cohan}]{beltagy2019scibert}
Iz~Beltagy, Kyle Lo, and Arman Cohan. 2019.
\newblock Scibert: A pretrained language model for scientific text.
\newblock \emph{arXiv preprint arXiv:1903.10676}.

\bibitem[{Bolukbasi et~al.(2016)Bolukbasi, Chang, Zou, Saligrama, and
  Kalai}]{bolukbasi2016man}
Tolga Bolukbasi, Kai-Wei Chang, James Zou, Venkatesh Saligrama, and Adam Kalai.
  2016.
\newblock \href {http://arxiv.org/abs/1607.06520} {Man is to computer
  programmer as woman is to homemaker? debiasing word embeddings}.

\bibitem[{Bornmann and Mutz(2015)}]{bornmann2015growth}
Lutz Bornmann and R{\"u}diger Mutz. 2015.
\newblock Growth rates of modern science: A bibliometric analysis based on the
  number of publications and cited references.
\newblock \emph{Journal of the Association for Information Science and
  Technology}, 66(11):2215--2222.

\bibitem[{Brown et~al.(2020)Brown, Mann, Ryder, Subbiah, Kaplan, Dhariwal,
  Neelakantan, Shyam, Sastry, Askell et~al.}]{brown2020language}
Tom~B Brown, Benjamin Mann, Nick Ryder, Melanie Subbiah, Jared Kaplan, Prafulla
  Dhariwal, Arvind Neelakantan, Pranav Shyam, Girish Sastry, Amanda Askell,
  et~al. 2020.
\newblock Language models are few-shot learners.
\newblock \emph{arXiv preprint arXiv:2005.14165}.

\bibitem[{Cachola et~al.(2020{\natexlab{a}})Cachola, Lo, Cohan, and
  Weld}]{cachola-etal-2020-tldr}
Isabel Cachola, Kyle Lo, Arman Cohan, and Daniel Weld. 2020{\natexlab{a}}.
\newblock \href {https://doi.org/10.18653/v1/2020.findings-emnlp.428} {{TLDR}:
  Extreme summarization of scientific documents}.
\newblock In \emph{Findings of the Association for Computational Linguistics:
  EMNLP 2020}, pages 4766--4777, Online. Association for Computational
  Linguistics.

\bibitem[{Cachola et~al.(2020{\natexlab{b}})Cachola, Lo, Cohan, and
  Weld}]{Cachola2020TLDRES}
Isabel Cachola, Kyle Lo, Arman Cohan, and Daniel~S. Weld. 2020{\natexlab{b}}.
\newblock Tldr: Extreme summarization of scientific documents.
\newblock \emph{ArXiv}, abs/2004.15011.

\bibitem[{Chakraborty et~al.(2020)Chakraborty, Goyal, and
  Mukherjee}]{chakraborty2020aspect}
Souvic Chakraborty, Pawan Goyal, and Animesh Mukherjee. 2020.
\newblock Aspect-based sentiment analysis of scientific reviews.
\newblock \emph{arXiv preprint arXiv:2006.03257}.

\bibitem[{Chen and Bansal(2018)}]{chen2018fast}
Yen-Chun Chen and Mohit Bansal. 2018.
\newblock Fast abstractive summarization with reinforce-selected sentence
  rewriting.
\newblock In \emph{Proceedings of the 56th Annual Meeting of the Association
  for Computational Linguistics (Volume 1: Long Papers)}, volume~1, pages
  675--686.

\bibitem[{Cohan et~al.(2018{\natexlab{a}})Cohan, Dernoncourt, Kim, Bui, Kim,
  Chang, and Goharian}]{Cohan2018ADA}
Arman Cohan, Franck Dernoncourt, Doo~Soon Kim, Trung Bui, Seokhwan Kim,
  W.~Chang, and Nazli Goharian. 2018{\natexlab{a}}.
\newblock A discourse-aware attention model for abstractive summarization of
  long documents.
\newblock In \emph{NAACL-HLT}.

\bibitem[{Cohan et~al.(2018{\natexlab{b}})Cohan, Dernoncourt, Kim, Bui, Kim,
  Chang, and Goharian}]{cohan-etal-2018-discourse}
Arman Cohan, Franck Dernoncourt, Doo~Soon Kim, Trung Bui, Seokhwan Kim, Walter
  Chang, and Nazli Goharian. 2018{\natexlab{b}}.
\newblock \href {https://doi.org/10.18653/v1/N18-2097} {A discourse-aware
  attention model for abstractive summarization of long documents}.
\newblock In \emph{Proceedings of the 2018 Conference of the North {A}merican
  Chapter of the Association for Computational Linguistics: Human Language
  Technologies, Volume 2 (Short Papers)}, pages 615--621, New Orleans,
  Louisiana. Association for Computational Linguistics.

\bibitem[{Cohan and Goharian(2017)}]{cohan2017scientific}
Arman Cohan and Nazli Goharian. 2017.
\newblock Scientific article summarization using citation-context and article's
  discourse structure.
\newblock \emph{arXiv preprint arXiv:1704.06619}.

\bibitem[{De~Bellis(2009)}]{de2009bibliometrics}
Nicola De~Bellis. 2009.
\newblock \emph{Bibliometrics and citation analysis: from the science citation
  index to cybermetrics}.
\newblock scarecrow press.

\bibitem[{Devlin et~al.(2019)Devlin, Chang, Lee, and
  Toutanova}]{devlin2019bert}
Jacob Devlin, Ming-Wei Chang, Kenton Lee, and Kristina Toutanova. 2019.
\newblock Bert: Pre-training of deep bidirectional transformers for language
  understanding.
\newblock In \emph{Proceedings of the 2019 Conference of the North American
  Chapter of the Association for Computational Linguistics: Human Language
  Technologies, Volume 1 (Long and Short Papers)}, pages 4171--4186.

\bibitem[{Dou et~al.(2020)Dou, Liu, Hayashi, Jiang, and Neubig}]{dou2020gsum}
Zi-Yi Dou, Pengfei Liu, Hiroaki Hayashi, Zhengbao Jiang, and Graham Neubig.
  2020.
\newblock Gsum: A general framework for guided neural abstractive
  summarization.
\newblock \emph{arXiv preprint arXiv:2010.08014}.

\bibitem[{Efron(1992)}]{efron1992bootstrap}
Bradley Efron. 1992.
\newblock Bootstrap methods: another look at the jackknife.
\newblock In \emph{Breakthroughs in statistics}, pages 569--593. Springer.

\bibitem[{Erera et~al.(2019)Erera, Shmueli-Scheuer, Feigenblat, Nakash, Boni,
  Roitman, Cohen, Weiner, Mass, Rivlin, Lev, Jerbi, Herzig, Hou, Jochim,
  Gleize, Bonin, and Konopnicki}]{Erera2019ASS}
Shai Erera, Michal Shmueli-Scheuer, Guy Feigenblat, O.~Nakash, O.~Boni, Haggai
  Roitman, Doron Cohen, B.~Weiner, Y.~Mass, Or~Rivlin, G.~Lev, Achiya Jerbi,
  Jonathan Herzig, Yufang Hou, Charles Jochim, Martin Gleize, F.~Bonin, and
  D.~Konopnicki. 2019.
\newblock A summarization system for scientific documents.
\newblock In \emph{EMNLP/IJCNLP}.

\bibitem[{Feigenblat et~al.(2017)Feigenblat, Roitman, Boni, and
  Konopnicki}]{10.1145/3077136.3080690}
Guy Feigenblat, Haggai Roitman, Odellia Boni, and David Konopnicki. 2017.
\newblock \href {https://doi.org/10.1145/3077136.3080690} {Unsupervised
  query-focused multi-document summarization using the cross entropy method}.
\newblock In \emph{Proceedings of the 40th International ACM SIGIR Conference
  on Research and Development in Information Retrieval}, SIGIR '17, page
  961–964, New York, NY, USA. Association for Computing Machinery.

\bibitem[{Frermann and Klementiev(2019)}]{frermann-klementiev-2019-inducing}
Lea Frermann and Alexandre Klementiev. 2019.
\newblock \href {https://doi.org/10.18653/v1/P19-1630} {Inducing document
  structure for aspect-based summarization}.
\newblock In \emph{Proceedings of the 57th Annual Meeting of the Association
  for Computational Linguistics}, pages 6263--6273, Florence, Italy.
  Association for Computational Linguistics.

\bibitem[{Gao et~al.(2019)Gao, Eger, Kuznetsov, Gurevych, and
  Miyao}]{gao-etal-2019-rebuttal}
Yang Gao, Steffen Eger, Ilia Kuznetsov, Iryna Gurevych, and Yusuke Miyao. 2019.
\newblock \href {https://doi.org/10.18653/v1/N19-1129} {Does my rebuttal
  matter? insights from a major {NLP} conference}.
\newblock In \emph{Proceedings of the 2019 Conference of the North {A}merican
  Chapter of the Association for Computational Linguistics: Human Language
  Technologies, Volume 1 (Long and Short Papers)}, pages 1274--1290,
  Minneapolis, Minnesota. Association for Computational Linguistics.

\bibitem[{Gehrmann et~al.(2018)Gehrmann, Deng, and Rush}]{gehrmann2018bottom}
Sebastian Gehrmann, Yuntian Deng, and Alexander Rush. 2018.
\newblock Bottom-up abstractive summarization.
\newblock In \emph{Proceedings of the 2018 Conference on Empirical Methods in
  Natural Language Processing}, pages 4098--4109.

\bibitem[{Hayashi et~al.(2020)Hayashi, Budania, Wang, Ackerson, Neervannan, and
  Neubig}]{hayashi20tacl}
Hiroaki Hayashi, Prashant Budania, Peng Wang, Chris Ackerson, Raj Neervannan,
  and Graham Neubig. 2020.
\newblock \href {https://arxiv.org/abs/2011.07832} {Wikiasp: A dataset for
  multi-domain aspect-based summarization}.
\newblock \emph{Transactions of the Association for Computational Linguistics
  (TACL)}.

\bibitem[{He et~al.(2016)He, Zhang, Ren, and Sun}]{he2016deep}
Kaiming He, Xiangyu Zhang, Shaoqing Ren, and Jian Sun. 2016.
\newblock Deep residual learning for image recognition.
\newblock In \emph{Proceedings of the IEEE conference on computer vision and
  pattern recognition}, pages 770--778.

\bibitem[{Hermann et~al.(2015)Hermann, Kocisky, Grefenstette, Espeholt, Kay,
  Suleyman, and Blunsom}]{hermann2015teaching}
Karl~Moritz Hermann, Tomas Kocisky, Edward Grefenstette, Lasse Espeholt, Will
  Kay, Mustafa Suleyman, and Phil Blunsom. 2015.
\newblock Teaching machines to read and comprehend.
\newblock In \emph{Advances in Neural Information Processing Systems}, pages
  1684--1692.

\bibitem[{Hou et~al.(2019)Hou, Jochim, Gleize, Bonin, and
  Ganguly}]{hou-etal-2019-identification}
Yufang Hou, Charles Jochim, Martin Gleize, Francesca Bonin, and Debasis
  Ganguly. 2019.
\newblock \href {https://doi.org/10.18653/v1/P19-1513} {Identification of
  tasks, datasets, evaluation metrics, and numeric scores for scientific
  leaderboards construction}.
\newblock In \emph{Proceedings of the 57th Annual Meeting of the Association
  for Computational Linguistics}, pages 5203--5213, Florence, Italy.
  Association for Computational Linguistics.

\bibitem[{Huang(2018)}]{huang2018deep}
Jia-Bin Huang. 2018.
\newblock Deep paper gestalt.
\newblock \emph{arXiv preprint arXiv:1812.08775}.

\bibitem[{Jain et~al.(2020)Jain, van Zuylen, Hajishirzi, and
  Beltagy}]{jain-etal-2020-scirex}
Sarthak Jain, Madeleine van Zuylen, Hannaneh Hajishirzi, and Iz~Beltagy. 2020.
\newblock \href {https://doi.org/10.18653/v1/2020.acl-main.670} {{S}ci{REX}:
  {A} challenge dataset for document-level information extraction}.
\newblock In \emph{Proceedings of the 58th Annual Meeting of the Association
  for Computational Linguistics}, pages 7506--7516, Online. Association for
  Computational Linguistics.

\bibitem[{Jecmen et~al.(2020)Jecmen, Zhang, Liu, Shah, Conitzer, and
  Fang}]{jecmen2020mitigating}
Steven Jecmen, Hanrui Zhang, Ryan Liu, Nihar~B Shah, Vincent Conitzer, and Fei
  Fang. 2020.
\newblock Mitigating manipulation in peer review via randomized reviewer
  assignments.
\newblock \emph{arXiv preprint arXiv:2006.16437}.

\bibitem[{Jefferson et~al.(2002{\natexlab{a}})Jefferson, Alderson, Wager, and
  Davidoff}]{jefferson2002effects}
Tom Jefferson, Philip Alderson, Elizabeth Wager, and Frank Davidoff.
  2002{\natexlab{a}}.
\newblock Effects of editorial peer review: a systematic review.
\newblock \emph{Jama}, 287(21):2784--2786.

\bibitem[{Jefferson et~al.(2002{\natexlab{b}})Jefferson, Wager, and
  Davidoff}]{jefferson2002measuring}
Tom Jefferson, Elizabeth Wager, and Frank Davidoff. 2002{\natexlab{b}}.
\newblock Measuring the quality of editorial peer review.
\newblock \emph{Jama}, 287(21):2786--2790.

\bibitem[{Jha et~al.(2013)Jha, Abu-Jbara, and Radev}]{jha-etal-2013-system}
Rahul Jha, Amjad Abu-Jbara, and Dragomir Radev. 2013.
\newblock \href {https://www.aclweb.org/anthology/P13-2102} {A system for
  summarizing scientific topics starting from keywords}.
\newblock In \emph{Proceedings of the 51st Annual Meeting of the Association
  for Computational Linguistics (Volume 2: Short Papers)}, pages 572--577,
  Sofia, Bulgaria. Association for Computational Linguistics.

\bibitem[{Jha et~al.(2015{\natexlab{a}})Jha, Coke, and
  Radev}]{DBLP:conf/aaai/JhaCR15}
Rahul Jha, Reed Coke, and Dragomir~R. Radev. 2015{\natexlab{a}}.
\newblock \href {http://www.aaai.org/ocs/index.php/AAAI/AAAI15/paper/view/9855}
  {Surveyor: {A} system for generating coherent survey articles for scientific
  topics}.
\newblock In \emph{Proceedings of the Twenty-Ninth {AAAI} Conference on
  Artificial Intelligence, January 25-30, 2015, Austin, Texas, {USA}}, pages
  2167--2173. {AAAI} Press.

\bibitem[{Jha et~al.(2015{\natexlab{b}})Jha, Finegan-Dollak, King, Coke, and
  Radev}]{jha-etal-2015-content}
Rahul Jha, Catherine Finegan-Dollak, Ben King, Reed Coke, and Dragomir Radev.
  2015{\natexlab{b}}.
\newblock \href {https://doi.org/10.3115/v1/P15-1043} {Content models for
  survey generation: A factoid-based evaluation}.
\newblock In \emph{Proceedings of the 53rd Annual Meeting of the Association
  for Computational Linguistics and the 7th International Joint Conference on
  Natural Language Processing (Volume 1: Long Papers)}, pages 441--450,
  Beijing, China. Association for Computational Linguistics.

\bibitem[{Jin et~al.(2017)Jin, Geng, Zhao, and Zhang}]{jin2017integrating}
Jian Jin, Qian Geng, Qian Zhao, and Lixue Zhang. 2017.
\newblock Integrating the trend of research interest for reviewer assignment.
\newblock In \emph{Proceedings of the 26th International Conference on World
  Wide Web Companion}, pages 1233--1241.

\bibitem[{Kang et~al.(2018)Kang, Ammar, Dalvi, van Zuylen, Kohlmeier, Hovy, and
  Schwartz}]{kang18naacl}
Dongyeop Kang, Waleed Ammar, Bhavana Dalvi, Madeleine van Zuylen, Sebastian
  Kohlmeier, Eduard Hovy, and Roy Schwartz. 2018.
\newblock \href {https://arxiv.org/abs/1804.09635} {A dataset of peer reviews
  (peerread): Collection, insights and nlp applications}.
\newblock In \emph{Meeting of the North American Chapter of the Association for
  Computational Linguistics (NAACL)}, New Orleans, USA.

\bibitem[{Kingma and Ba(2014)}]{kingma2014adam}
Diederik Kingma and Jimmy Ba. 2014.
\newblock Adam: A method for stochastic optimization.
\newblock \emph{arXiv preprint arXiv:1412.6980}.

\bibitem[{Langford and Guzdial(2015)}]{langford2015arbitrariness}
John Langford and Mark Guzdial. 2015.
\newblock The arbitrariness of reviews, and advice for school administrators.

\bibitem[{Lewis et~al.(2019)Lewis, Liu, Goyal, Ghazvininejad, Mohamed, Levy,
  Stoyanov, and Zettlemoyer}]{Lewis2019BARTDS}
Mike Lewis, Yinhan Liu, Naman Goyal, Marjan Ghazvininejad, Abdelrahman Mohamed,
  Omer Levy, Ves Stoyanov, and Luke Zettlemoyer. 2019.
\newblock Bart: Denoising sequence-to-sequence pre-training for natural
  language generation, translation, and comprehension.
\newblock \emph{ArXiv}, abs/1910.13461.

\bibitem[{Lin and Hovy(2003)}]{lin2003automatic}
Chin-Yew Lin and Eduard Hovy. 2003.
\newblock Automatic evaluation of summaries using n-gram co-occurrence
  statistics.
\newblock In \emph{Proceedings of the 2003 Human Language Technology Conference
  of the North American Chapter of the Association for Computational
  Linguistics}, pages 150--157.

\bibitem[{Lo et~al.(2020)Lo, Wang, Neumann, Kinney, and
  Weld}]{lo-wang-2020-s2orc}
Kyle Lo, Lucy~Lu Wang, Mark Neumann, Rodney Kinney, and Daniel Weld. 2020.
\newblock \href {https://doi.org/10.18653/v1/2020.acl-main.447} {{S}2{ORC}: The
  semantic scholar open research corpus}.
\newblock In \emph{Proceedings of the 58th Annual Meeting of the Association
  for Computational Linguistics}, pages 4969--4983, Online. Association for
  Computational Linguistics.

\bibitem[{Luan et~al.(2018)Luan, He, Ostendorf, and Hajishirzi}]{luan2018multi}
Yi~Luan, Luheng He, Mari Ostendorf, and Hannaneh Hajishirzi. 2018.
\newblock Multi-task identification of entities, relations, and coreference for
  scientific knowledge graph construction.
\newblock \emph{arXiv preprint arXiv:1808.09602}.

\bibitem[{Manzoor and Shah(2020)}]{manzoor2020uncovering}
Emaad Manzoor and Nihar~B. Shah. 2020.
\newblock \href {http://arxiv.org/abs/2010.15300} {Uncovering latent biases in
  text: Method and application to peer review}.

\bibitem[{Mohammad et~al.(2009)Mohammad, Dorr, Egan, Hassan, Muthukrishan,
  Qazvinian, Radev, and Zajic}]{mohammad-etal-2009-using}
Saif Mohammad, Bonnie Dorr, Melissa Egan, Ahmed Hassan, Pradeep Muthukrishan,
  Vahed Qazvinian, Dragomir Radev, and David Zajic. 2009.
\newblock \href {https://www.aclweb.org/anthology/N09-1066} {Using citations to
  generate surveys of scientific paradigms}.
\newblock In \emph{Proceedings of Human Language Technologies: The 2009 Annual
  Conference of the North {A}merican Chapter of the Association for
  Computational Linguistics}, pages 584--592, Boulder, Colorado. Association
  for Computational Linguistics.

\bibitem[{Nallapati et~al.(2017)Nallapati, Zhai, and
  Zhou}]{Nallapati2017SummaRuNNerAR}
Ramesh Nallapati, Feifei Zhai, and Bowen Zhou. 2017.
\newblock Summarunner: A recurrent neural network based sequence model for
  extractive summarization of documents.
\newblock \emph{ArXiv}, abs/1611.04230.

\bibitem[{Narayan et~al.(2018)Narayan, Cohen, and Lapata}]{xsum-emnlp}
Shashi Narayan, Shay~B. Cohen, and Mirella Lapata. 2018.
\newblock Don't give me the details, just the summary! {T}opic-aware
  convolutional neural networks for extreme summarization.
\newblock In \emph{Proceedings of the 2018 Conference on Empirical Methods in
  Natural Language Processing}, Brussels, Belgium.

\bibitem[{Nguyen et~al.(2018)Nguyen, S{\'a}nchez-Hern{\'a}ndez, Agell, Rovira,
  and Angulo}]{nguyen2018decision}
Jennifer Nguyen, Germ{\'a}n S{\'a}nchez-Hern{\'a}ndez, N{\'u}ria Agell, Xari
  Rovira, and Cecilio Angulo. 2018.
\newblock A decision support tool using order weighted averaging for conference
  review assignment.
\newblock \emph{Pattern Recognition Letters}, 105:114--120.

\bibitem[{Paulus et~al.(2017)Paulus, Xiong, and Socher}]{paulus2017deep}
Romain Paulus, Caiming Xiong, and Richard Socher. 2017.
\newblock A deep reinforced model for abstractive summarization.
\newblock \emph{arXiv preprint arXiv:1705.04304}.

\bibitem[{Qazvinian and Radev(2008)}]{10.5555/1599081.1599168}
Vahed Qazvinian and Dragomir~R. Radev. 2008.
\newblock Scientific paper summarization using citation summary networks.
\newblock In \emph{Proceedings of the 22nd International Conference on
  Computational Linguistics - Volume 1}, COLING '08, page 689–696, USA.
  Association for Computational Linguistics.

\bibitem[{Qiao et~al.(2018)Qiao, Xu, and Han}]{qiao2018modularized}
Feng Qiao, Lizhen Xu, and Xiaowei Han. 2018.
\newblock Modularized and attention-based recurrent convolutional neural
  network for automatic academic paper aspect scoring.
\newblock In \emph{International Conference on Web Information Systems and
  Applications}, pages 68--76. Springer.

\bibitem[{Radford et~al.(2019)Radford, Wu, Child, Luan, Amodei, and
  Sutskever}]{radford2019language}
Alec Radford, Jeffrey Wu, Rewon Child, David Luan, Dario Amodei, and Ilya
  Sutskever. 2019.
\newblock Language models are unsupervised multitask learners.
\newblock \emph{OpenAI blog}, 1(8):9.

\bibitem[{Rae et~al.(2020)Rae, Potapenko, Jayakumar, and
  Lillicrap}]{Rae2020CompressiveTF}
Jack~W. Rae, Anna Potapenko, Siddhant~M. Jayakumar, and T.~Lillicrap. 2020.
\newblock Compressive transformers for long-range sequence modelling.
\newblock \emph{ArXiv}, abs/1911.05507.

\bibitem[{Rogers and Augenstein(2020)}]{rogers-augenstein-2020-improve}
Anna Rogers and Isabelle Augenstein. 2020.
\newblock \href {https://doi.org/10.18653/v1/2020.findings-emnlp.112} {What can
  we do to improve peer review in {NLP}?}
\newblock In \emph{Findings of the Association for Computational Linguistics:
  EMNLP 2020}, pages 1256--1262, Online. Association for Computational
  Linguistics.

\bibitem[{Rubinstein and Kroese(2013)}]{rubinstein2013cross}
Reuven~Y Rubinstein and Dirk~P Kroese. 2013.
\newblock \emph{The cross-entropy method: a unified approach to combinatorial
  optimization, Monte-Carlo simulation and machine learning}.
\newblock Springer Science \& Business Media.

\bibitem[{Smith(2006)}]{Smith2006PeerRA}
R.~Smith. 2006.
\newblock Peer review: A flawed process at the heart of science and journals.
\newblock \emph{Journal of the Royal Society of Medicine}, 99:178 -- 182.

\bibitem[{Stanovsky et~al.(2019)Stanovsky, Smith, and
  Zettlemoyer}]{stanovsky-etal-2019-evaluating}
Gabriel Stanovsky, Noah~A. Smith, and Luke Zettlemoyer. 2019.
\newblock \href {https://doi.org/10.18653/v1/P19-1164} {Evaluating gender bias
  in machine translation}.
\newblock In \emph{Proceedings of the 57th Annual Meeting of the Association
  for Computational Linguistics}, pages 1679--1684, Florence, Italy.
  Association for Computational Linguistics.

\bibitem[{Stelmakh et~al.(2019)Stelmakh, Shah, and Singh}]{stelmakh2019testing}
Ivan Stelmakh, Nihar Shah, and Aarti Singh. 2019.
\newblock On testing for biases in peer review.
\newblock In \emph{Advances in Neural Information Processing Systems}, pages
  5286--5296.

\bibitem[{Subramanian et~al.(2019)Subramanian, Li, Pilault, and
  Pal}]{subramanian2019extractive}
Sandeep Subramanian, Raymond Li, Jonathan Pilault, and Christopher Pal. 2019.
\newblock On extractive and abstractive neural document summarization with
  transformer language models.
\newblock \emph{arXiv preprint arXiv:1909.03186}.

\bibitem[{Tabah(1999)}]{tabah1999literature}
Albert~N Tabah. 1999.
\newblock Literature dynamics: Studies on growth, diffusion, and epidemics.
\newblock \emph{Annual review of information science and technology (ARIST)},
  34:249--86.

\bibitem[{Tomkins et~al.(2017)Tomkins, Zhang, and
  Heavlin}]{tomkins2017reviewer}
Andrew Tomkins, Min Zhang, and William~D Heavlin. 2017.
\newblock Reviewer bias in single-versus double-blind peer review.
\newblock \emph{Proceedings of the National Academy of Sciences},
  114(48):12708--12713.

\bibitem[{Vaswani et~al.(2017)Vaswani, Shazeer, Parmar, Uszkoreit, Jones,
  Gomez, Kaiser, and Polosukhin}]{vaswani2017attention}
Ashish Vaswani, Noam Shazeer, Niki Parmar, Jakob Uszkoreit, Llion Jones,
  Aidan~N Gomez, {\L}ukasz Kaiser, and Illia Polosukhin. 2017.
\newblock Attention is all you need.
\newblock In \emph{Advances in neural information processing systems}, pages
  5998--6008.

\bibitem[{Von~Bearnensquash(2010)}]{von2010paper}
Carven Von~Bearnensquash. 2010.
\newblock Paper gestalt.
\newblock \emph{Secret Proceedings of Computer Vision and Pattern Recognition
  (CVPR)}.

\bibitem[{Wadden et~al.(2020)Wadden, Lin, Lo, Wang, van Zuylen, Cohan, and
  Hajishirzi}]{wadden-etal-2020-fact}
David Wadden, Shanchuan Lin, Kyle Lo, Lucy~Lu Wang, Madeleine van Zuylen, Arman
  Cohan, and Hannaneh Hajishirzi. 2020.
\newblock \href {https://doi.org/10.18653/v1/2020.emnlp-main.609} {Fact or
  fiction: Verifying scientific claims}.
\newblock In \emph{Proceedings of the 2020 Conference on Empirical Methods in
  Natural Language Processing (EMNLP)}, pages 7534--7550, Online. Association
  for Computational Linguistics.

\bibitem[{Wang and Wan(2018)}]{wang2018sentiment}
Ke~Wang and Xiaojun Wan. 2018.
\newblock Sentiment analysis of peer review texts for scholarly papers.
\newblock In \emph{The 41st International ACM SIGIR Conference on Research \&
  Development in Information Retrieval}, pages 175--184.

\bibitem[{Wang et~al.(2020)Wang, Zeng, Huang, Knight, Ji, and
  Rajani}]{wang2020reviewrobot}
Qingyun Wang, Qi~Zeng, Lifu Huang, Kevin Knight, Heng Ji, and Nazneen~Fatema
  Rajani. 2020.
\newblock Reviewrobot: Explainable paper review generation based on knowledge
  synthesis.
\newblock In \emph{Proceedings of INLG}.

\bibitem[{Xiao and Carenini(2019)}]{Xiao2019ExtractiveSO}
Wen Xiao and Giuseppe Carenini. 2019.
\newblock Extractive summarization of long documents by combining global and
  local context.
\newblock \emph{ArXiv}, abs/1909.08089.

\bibitem[{Xing et~al.(2020)Xing, Fan, and Wan}]{xing-etal-2020-automatic}
Xinyu Xing, Xiaosheng Fan, and Xiaojun Wan. 2020.
\newblock \href {https://doi.org/10.18653/v1/2020.acl-main.550} {Automatic
  generation of citation texts in scholarly papers: A pilot study}.
\newblock In \emph{Proceedings of the 58th Annual Meeting of the Association
  for Computational Linguistics}, pages 6181--6190, Online. Association for
  Computational Linguistics.

\bibitem[{Xiong and Litman(2011)}]{xiong-litman-2011-automatically}
Wenting Xiong and Diane Litman. 2011.
\newblock \href {https://www.aclweb.org/anthology/P11-2088} {Automatically
  predicting peer-review helpfulness}.
\newblock In \emph{Proceedings of the 49th Annual Meeting of the Association
  for Computational Linguistics: Human Language Technologies}, pages 502--507,
  Portland, Oregon, USA. Association for Computational Linguistics.

\bibitem[{Yasunaga et~al.(2019{\natexlab{a}})Yasunaga, Kasai, Zhang, Fabbri,
  Li, Friedman, and Radev}]{Yasunaga2019ScisummNetAL}
Michihiro Yasunaga, Jungo Kasai, Rui Zhang, A.~R. Fabbri, Irene Li,
  D.~Friedman, and Dragomir~R. Radev. 2019{\natexlab{a}}.
\newblock Scisummnet: A large annotated corpus and content-impact models for
  scientific paper summarization with citation networks.
\newblock In \emph{AAAI}.

\bibitem[{Yasunaga et~al.(2019{\natexlab{b}})Yasunaga, Kasai, Zhang, Fabbri,
  Li, Friedman, and Radev}]{DBLP:conf/aaai/YasunagaKZFLFR19}
Michihiro Yasunaga, Jungo Kasai, Rui Zhang, Alexander~R. Fabbri, Irene Li, Dan
  Friedman, and Dragomir~R. Radev. 2019{\natexlab{b}}.
\newblock \href {https://doi.org/10.1609/aaai.v33i01.33017386} {Scisummnet: {A}
  large annotated corpus and content-impact models for scientific paper
  summarization with citation networks}.
\newblock In \emph{The Thirty-Third {AAAI} Conference on Artificial
  Intelligence, {AAAI} 2019, The Thirty-First Innovative Applications of
  Artificial Intelligence Conference, {IAAI} 2019, The Ninth {AAAI} Symposium
  on Educational Advances in Artificial Intelligence, {EAAI} 2019, Honolulu,
  Hawaii, USA, January 27 - February 1, 2019}, pages 7386--7393. {AAAI} Press.

\bibitem[{Zhang et~al.(2019)Zhang, Kishore, Wu, Weinberger, and
  Artzi}]{zhang2019bertscore}
Tianyi Zhang, Varsha Kishore, Felix Wu, Kilian~Q Weinberger, and Yoav Artzi.
  2019.
\newblock Bertscore: Evaluating text generation with bert.
\newblock \emph{arXiv}, pages arXiv--1904.

\bibitem[{Zhao et~al.(2018)Zhao, Wang, Yatskar, Ordonez, and
  Chang}]{zhao2018gender}
Jieyu Zhao, Tianlu Wang, Mark Yatskar, Vicente Ordonez, and Kai-Wei Chang.
  2018.
\newblock Gender bias in coreference resolution: Evaluation and debiasing
  methods.
\newblock \emph{arXiv preprint arXiv:1804.06876}.

\end{thebibliography}
\bibliographystyle{acl_natbib}

\newpage
\appendix

\section{Appendices}
\label{sec:appendix}

\subsection{Reviews of this Paper Written by Our Model} \label{sec:self-review}

Notably, the following review is generated based on the paper without this review as well as the TL;QR section (The original version can be found here: \url{https://drive.google.com/file/d/1nC4kCaaeqKRiajcvK75g421Ku9Jog1n9/view?usp=sharing}). And we directly put the system output here without any grammar check.

\noindent \textbf{Summary} : This paper presents an approach to evaluate the quality of reviews generated by an automatic summarization system for scientific papers . The authors build a dataset of reviews , named ASAP-Review1 , from machine learning domain , and make fine-grained annotations of aspect information for each review , which provides the possibility for a richer evaluation of generated reviews . They train a summarization model to generate reviews from scientific papers , and evaluate the output according to our evaluation metrics described above . They propose different architectural designs for this model , which they dub ReviewAdvisor , and comprehensively evaluate them , interpreting their relative advantages and disadvantages . They find that both human and automatic reviewers exhibit varying degrees of bias regarding native English speakers vs non-native English speakers , and find that native speakers tend to obtain higher scores on “ Clarity ” and “ Potential Impact ” . The paper is well-written and easy to follow .

\noindent \textbf{Strengths} : 1 . The proposed approach is novel and interesting . 2 .The paper is easy to read and well-organized . 3 .The evaluation metrics are well-motivated . 4 .The authors have done a good job of evaluating the proposed approach .

\noindent \textbf{Weaknesses} : 1 ) The evaluation metrics used in this paper are not well-defined . For example , what is the definition of “ good review quality ” ? What is the criteria for a good review ? 2 ) It is not clear to me how the authors define “ factually correct ” , “ fair ” or “ non-factual ” in Section 3.2 . 3 ) The authors should provide more details about the evaluation metrics in the paper . For instance , what are the criteria used in Table 1 and Table 2 ? What are the metrics used for the evaluation in Table 3 and Table 4 ? 4 ) It would be better if the authors can provide more explanations about the results of Table 2 and Table 3 . 5 ) In Table 3 , the authors mentioned that “ we found the constructed automatic review system generates nonfactual statements regarding many aspects of the paper assess- 1ASpect-enhanced-Anced Peer Review dataset , which is a serious flaw in a high-stakes setting such as reviewing . However , there are some bright points as well . ” However , it is unclear to me why the authors found this problem . 6 ) In Section 4.3 , it seems that the authors did not provide any explanation about why the human reviewers are biased . 7 ) In Figure 1 , it would be good to provide more information about the training data . 8 ) In section 4.4 , it will be better to provide some explanation about how the human reviews are generated .

\subsection{Screenshot of Our Demo System} \label{sec:screenshot}
\begin{figure*}
    \centering
    \includegraphics[width=1\linewidth]{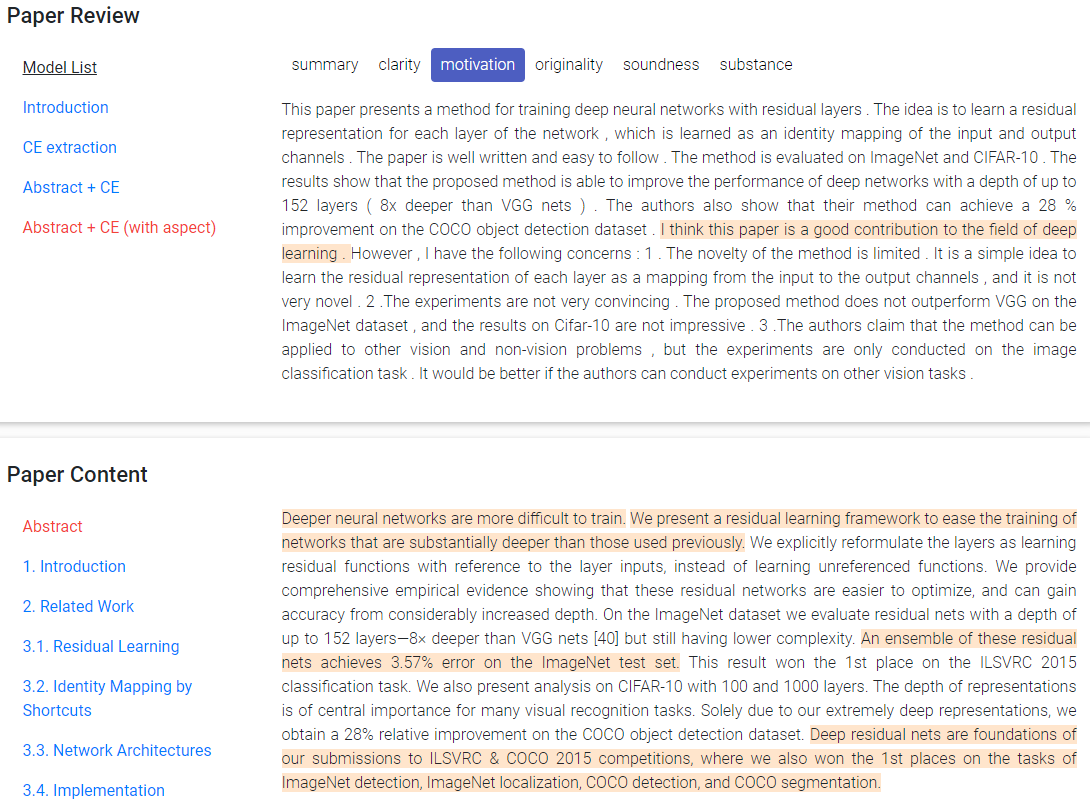}
    \caption{Generated reviews with aspects by our \textit{ReviewAdvisor} demo associated with evidence sentences from the paper ``Deep Residual Learning for Image Recognition" \cite{he2016deep}.}
    \label{fig:my_label}
\end{figure*}

\subsection{Details for Evaluation Metrics}
\label{app:detail-metric}

\paragraph{\textsc{Rec}}In \S\ref{sec: multi-pers-eval}, the \textsc{Rec} function we define corresponds to the recommendation sentiment of a review, with $\{-1, 0, 1\}$ representing negative, neutral and positive.

To decide the sentiment of a reference review, we use the rating information from reviewers: (i) if the rating corresponds to marginal accept or marginal reject, then we regard it as neutral; (ii) if the rating is above marginal accept, then we regard it as positive; (iii) otherwise, we regard it as negative.

To decide the sentiment of a generated review, two members of the project team judged the sentiment polarity of a review. If they agreed with each other, then we uses the agreed-upon sentiment, if they disagreed with each other, then we label the sentiment of that review as neutral. The Cohen kappa of two annotators is 0.5778, which is commonly referred to as ``moderate'' agreement.

\paragraph{\textsc{Info}} The judgement of evidence for negative aspects was conducted by a member of the project team, who judged whether each identified negative aspect was accompanied by evidence irrespective of the correctness of the evidence. In other words, as long as there is a reason, we count that as an evidence.

\paragraph{\textsc{SAcc} \& \textsc{ACon}}The judgement of summary accuracy and valid support for negative aspects are performed by one of the first authors of the reviewed paper. Summary and each negative aspect in the review should be scored 0, 0.5 or 1 which represent agree, partially agree and disagree respectively. We provide authors with the following instructions:

\begin{quote}
``\texttt{We have created a Google doc for your paper, and you can score the summary as well as each aspect with its corresponding comments inside the red brackets.
"1" denotes agree,
"0.5" denotes partially agree,
"0" denotes disagree.
You only need to assign a score based on your judgment. For summary, agree means that you think it's factually correct. For each aspect, agree means that you think the strength/weakness the reviewer points out is reasonable or constructive.}''
\end{quote}

\paragraph{BERTScore}
Regarding BERTScore, we apply the same rescaling procedure following \citet{zhang2019bertscore}, which does not affect the ranking ability of BERTScore, but make the scores more discriminative.

\subsection{Training of Aspect Tagger}
\label{app:train-tagger}

We formulate the annotation process as a sequence labeling problem where the input is a sequence consisting of $n$ words $S = w_1, \cdots, w_n$, and the target is a sequence of tags one for each word $T = t_1, \cdots, t_n$. We aim to find a mapping $f$ such that $T = f(S)$ can convey reasonable aspect information in the input sequence.

We first segment each review into multiple sentences and consider each sentence as an individual training example.\footnote{We also tried using larger context such as paragraph, but found out the results less satisfying since the model identified fewer aspects.}
For a tokenized sequence contains $n$ tokens $(w_1, w_2, \cdots, w_n)$, we use BERT to get a contextualized representation for each token $(\bm{e_1}, \bm{e_2}, \cdots, \bm{e_n})$, where $\bm{e_i}$ represents the vector for $i$-th token.
Then those contextualized representations can be used as features for token classification:
\begin{equation*}
    \mathbf{p}_i = \mathrm{softmax}(\mathbf{W}\bm{e_i} + \mathbf{b})
\end{equation*}
where $\mathbf{W}$ and $\mathbf{b}$ are tunable parameters of the multilayer perceptron. $\mathbf{p}_i$ is a vector that represents the probability of token $i$ being assigned to different aspects.

We use the negative log likelihood of the correct labels as training loss:
\begin{equation*}
    \mathcal{L} = -\sum_{t \in \mathcal{T}}\log \mathbf{p}_{tj}
\end{equation*}
where $j$ is the label of token $t$, and $\mathcal{T}$ denotes all the tokens.

\begin{table*}[h]
\footnotesize
\renewcommand{\arraystretch}{1.2}
\begin{center}
\setlength\tabcolsep{7pt}
\begin{tabular}{cll}
\toprule
\textbf{Heuristics} & \multicolumn{1}{c}{\textbf{Before}} & \multicolumn{1}{c}{\textbf{After}} \\ \midrule

\multirow{4}{*}{1} & \underline{The authors present a method for learning Hami-} & \underline{The authors present a method for learning Hami-} \\
& \underline{ltonian functions}$_\texttt{[Summary]}$                             \underline{$\cdots$}$_\texttt{[O]}$ \underline{this is}$_\texttt{[Summary]}$  & \underline{ltonian functions $\cdots$ this is $\cdots$ that maps past o-} \\
& \underline{$\cdots$}$_\texttt{[O]}$ \underline{that maps past observations to a latent p,} & \underline{bservations to a latent p, q space in a VAE-like fa-} \\
& \underline{q space in a VAE-like fashion.}$_\texttt{[Summary]}$ & \underline{shion.}$_\texttt{[Summary]}$\\
\midrule
\multirow{5}{*}{2} & \underline{This paper proposes a new representation learn-} & \underline{This paper proposes a new representation learn-} \\
& \underline{ing model for graph optimization, Graph2Seq .} & \underline{ing model for graph optimization, Graph2Seq .} \\
 & $_\texttt{[Summary]}$ $\cdots$ \underline{the theorems are very interesting .} & $_\texttt{[Summary]}$ $\cdots$ \underline{the theorems are very interesting .} \\
 & $_\texttt{[Positive Originality]}$ $\cdots$ \underline{The performance of} & $_\texttt{[Positive Originality]}$ $\cdots$ \underline{The performance of} \\
 & \underline{Graph2Seq is remarkable.}$_\texttt{[Summary]}$ & \underline{Graph2Seq is remarkable.}$_\texttt{[O]}$\\

\midrule

\multirow{3}{*}{3} & \underline{The proposed idea is novel}$_\texttt{[Positive Originality]}$ & \underline{The proposed idea is novel}$_\texttt{[Positive Originality]}$ \\
& \underline{.}$_\texttt{[Positive Motivation]}$ \underline{The paper is well written} & \underline{.}$_\texttt{[O]}$ \underline{The paper is well written and easy to follow.} \\
& \underline{and easy to follow.}$_\texttt{[Positive Clarity]}$ & $_\texttt{[Positive Clarity]}$ \\

\midrule

\multirow{3}{*}{4} & \underline{The overall notion of}$_\texttt{[Positive Originality]}$ \underline{learn-} & \underline{The overall notion of learning a Hamiltonian net-} \\
& \underline{ning}$_\texttt{[O]}$
\underline{a Hamiltonian network directly is a great} & \underline{work directly is a great one.}$_\texttt{[Positive Originality]}$ \\
& \underline{one.}$_\texttt{[Positive Originality]}$ & \\

\midrule

\multirow{2}{*}{5} & \underline{It is}$_\texttt{[O]}$ \underline{clearly}$_\texttt{[Positive Clarity]}$ \underline{geared towards}& \underline{It is clearly geared towards DNN practitioners.}$_\texttt{[O]}$\\
 &  \underline{DNN practitioners.}$_\texttt{[O]}$ \\

\midrule

\multirow{3}{*}{6} & \underline{In contrast , this aspect}$_\texttt{[O]}$ \underline{is missing from other} & \underline{In contrast, this aspect is missing from other work} \\
& \underline{work on ML}$_\texttt{[Negative Meaningful Comparison]}$ \underline{for} & \underline{on ML for optimization.}$_\texttt{[Negative Meaningful Comp.]}$\\
& \underline{optimization.}$_\texttt{[O]}$ & \\

\midrule

\multirow{6}{*}{7} & \underline{The authors propose a novel approach to estim-}  & \underline{The authors propose a novel approach to estimate} \\
& \underline{ate unbalanced optimal transport between sam-} & \underline{unbalanced optimal transport between sampled m-} \\
& \underline{pled measures that scales well in the dimension} & \underline{easures that scales well in the dimension and in the} \\
& \underline{and in the number of samples $\cdots$ The effectiv-} & \underline{in the number of samples $\cdots$ The effectiveness of} \\
& \underline{eness of the approach}$_\texttt{[Summary]}$ \underline{is shown on so-} & \underline{the approach is shown on some tasks.}$_\texttt{[Summary]}$\\
& \underline{me tasks.}$_\texttt{[O]}$ \\

\bottomrule
\end{tabular}
\end{center}

\caption{\label{tab:heuristic-rules}Examples of seven heuristic rules used for refineing prediction results.}
\end{table*}

We used 900 annotated reviews for training and 100 for validation which is equivalent to using 16,543 training data and 1,700 validation data since we consider sentence as the basic individual training sample. The initial BERT checkpoint we used is ``\texttt{bert-large-cased}'' which is the large version of BERT with an uncased vocabulary. We used Adam optimizer \citep{kingma2014adam} with a learning rate of $5e^{-5}$ to finetune our model. We trained for 5 epochs and saved the model that achieved lowest loss on validation set as our aspect tagger.

\begin{table*}[h]
\small
\begin{center}
\renewcommand{\arraystretch}{1.6}
\begin{tabular}{p{0.95\textwidth}}
\toprule
\mycbox{sum} summary \qquad
\mycbox{ori} originality \qquad
\mycbox{cla} clarity  \qquad
\mycbox{cmp} meaningful comparison \qquad
\mycbox{mot} motivation \qquad
\mycbox{sub} substance
\\ \midrule
\setlength{\fboxsep}{0pt}\colorbox{sum}{This paper studies the graph embedding problem by using encoder-decoder method . The experimental study on real ne-}\\
\setlength{\fboxsep}{0pt}\colorbox{sum}{twork data sets show the features extracted by the proposed model is good forclassification .} Strong points of this paper: \\

1. \setlength{\fboxsep}{0pt}\colorbox{ori}{The idea of using the methods from natural language processing to graph mining is quite interesting .} 2. \setlength{\fboxsep}{0pt}\colorbox{cla}{The organiz-}\\

\setlength{\fboxsep}{0pt}\colorbox{cla}{ation of the paper is clear Weak points of this paper}:
1. \setlength{\fboxsep}{0pt}\colorbox{cmp}{Comparisons with state-of-art-methods ( Graph Kernels ) is mis-}\\\setlength{\fboxsep}{0pt}\colorbox{cmp}{sing} . \setlength{\fboxsep}{0pt}\colorbox{mot}{2. The problem is not well motivated}, are there any application of this . What is the difference from the graph kernel \\
methods ? \setlength{\fboxsep}{0pt}\colorbox{cmp}{The comparison with graph kernel is missing .}
\setlength{\fboxsep}{0pt}\colorbox{sub}{3. Need more experiment to demonstrate the power of their fe-}\\
\setlength{\fboxsep}{0pt}\colorbox{sub}{ature extraction methods .} ( Clustering,
Search, Prediction etc.) \setlength{\fboxsep}{0pt}\colorbox{cla}{4. Presentation of the paper is weak . There are lots of ty-}\\\setlength{\fboxsep}{0pt}\colorbox{cla}{pos and unclear statements. }\\
\bottomrule
\end{tabular}
\end{center}
\caption{\label{annotate-tab} An example of automatically labeled reviews.}
\end{table*}

\subsection{Heuristics for Refining Prediction Results}
\label{sec:heuristics}
The seven heuristic rules used for refining the prediction results are listed below. Examples of those rules are shown in Tab.~\ref{tab:heuristic-rules}.
\begin{enumerate}
    \item If there are no other tags (they are tagged as ``\texttt{O}'' which stands for \textit{Outside}) between two ``\texttt{summary}'' tags, then replace all tags between them with ``\texttt{summary}'' tag.
    \item If there are multiple discontinuous text spans tagged as ``\texttt{summary}'', we keep the first one and discard others.
    \item If the punctuation is separately tagged and is different from its neighbors, we replace its tag to ``\texttt{O}''.
    \item If two identical tags are separated by a single other tag, then replace this tag with its right neighbor's tag.
    \item If there exists a single token with a tag and its neighbors are ``\texttt{O}'', then replace this tag to `\texttt{O}''.
    \item For a ``\texttt{non-summary}'' ``\texttt{non-O}'' tag span, if its neighbors are ``\texttt{O}'' and the start/end of this span is not special symbol (for example, punctuations or other symbols that have 1 length), then we expand from its start/end until we meet other ``\texttt{non-O}'' tag or special symbol.
    \item If the ``\texttt{summary}'' span does not end with a period, then we truncate or extend it at most five words to make it ends with a period.
\end{enumerate}

\subsection{An Example of Automatically Annotated Reviews}
\label{sec:example-annotate}

Tab.~\ref{annotate-tab} illustrates an annotated review after using our trained aspect tagger and heuristic refining rules in Appendix \ref{sec:heuristics}.
Although here we do not add separate polarity tags to avoid visual burden, the polarity of each aspect the model predicts is correct.

\subsection{Calculation of Aspect Precision and Aspect Recall}
\label{app:asp-pre-and-asp-recall}
To measure aspect precision, we asked three annotators to decide whether each aspect span the model predicted is accurate. They were asked to delete a tagged span if they regarded it as inappropriate. We denote all prediction spans as $\mathcal{M}$, and the filtered spans from annotators as $\mathcal{F}_1$, $\mathcal{F}_2$ and $\mathcal{F}_3$. We represent $n_{\mathcal{S}}$ as the total number of text spans in $\mathcal{S}$.
Here we define correct spans as

\begin{equation*}
    \mathcal{C} = \{l | l \in \mathcal{F}_1, l \in \mathcal{F}_2, l \in \mathcal{F}_3 \}
\end{equation*}
The aspect precision is calculated using Formula \ref{precision}.
\begin{equation}
    \label{precision}
    \text{Precision} = \frac{n_{\mathcal{C}}}{n_{\mathcal{M}}}
\end{equation}

For measuring aspect recall, we asked three annotators to label aspect spans that they identified while the model ignored. We denote the additional labeled spans from one annotator as $\mathcal{A}$ where $\mathcal{A} = \{a_1, a_2,\cdots, a_{n_{\mathcal{A}}}\}$, $a_i$ represents a text span. We denote the additional labeled spans from other two annotators as $\mathcal{B}$ and $\mathcal{C}$.

We define common ignored spans for every two annotators as below. $|\cdot|$ denotes the number of tokens in a span and $\cap$ takes the intersect span between two spans.
\begin{align*}
    \mathcal{I}_1 &= \{ a_i \cap b_j| \frac{|a_i \cap b_j|}{\min{\{|a_i|, |b_j|\}}} > 0.5\} \\
    \mathcal{I}_2 &= \{ b_i \cap c_j| \frac{|b_i \cap c_j|}{\min{\{|b_i|, |c_j|\}}} > 0.5\} \\
    \mathcal{I}_3 &= \{ a_i \cap c_j| \frac{|a_i \cap c_j|}{\min{\{|a_i|, |c_j|\}}} > 0.5\}
\end{align*}

We also define common ignored spans for three annotators as below.
\begin{equation*}
    \mathcal{I} = \{ a_i \cap b_j \cap c_k | \frac{|a_i \cap b_j \cap c_k|}{\min{\{|a_i|, |b_j|, |c_k|\}}} > 0.3\}
\end{equation*}
where $a_i$, $b_j$, $c_k$ are text spans from $\mathcal{A}$, $\mathcal{B}$, $\mathcal{C}$ respectively.
We assume all the spans the model predicts are correct. Then we can calculate total number of spans using Formula \ref{eq:total_num}.
\begin{equation}
\label{eq:total_num}
\begin{aligned}
        n = &n_{\mathcal{M}} + n_{\mathcal{A}} + n_{\mathcal{B}} + n_{\mathcal{C}} -n_{\mathcal{I}_1} -n_{\mathcal{I}_2}-\\
    &n_{\mathcal{I}_3} + n_{\mathcal{I}}
\end{aligned}
\end{equation}

The aspect recall is calculated using Formula \ref{eq:recall}.

\begin{equation}
    \label{eq:recall}
    \text{Recall} = \frac{n_{\mathcal{M}}}{n}
\end{equation}

\subsection{Adjusting BART for Long Documents}
\label{app: bart4long}

\begin{table*}[h]
\small
\centering
\begin{tabular*}{0.85\textwidth}{llllllll}
\toprule
\multicolumn{8}{l}{\textsc{Keywords}} \\ \midrule
    find & prove & examine & address & suggest & baseline & optimize & outperform \\
    show & design & explore & analyze & achieve & maximize & efficient & generalize \\
    imply & reduce & propose & explain & perform & minimize & effective & understand  \\
    study & metric & observe & benefit & improve & increase & introduce & investigate \\
    bound & better & present & compare & dataset & decrease & interpret & demonstrate  \\
    apply & result & develop & measure & evaluate & discover & experiment &state-of-the-art \\ \bottomrule

\end{tabular*}
    \caption{\label{table: keywords}Predefined keywords for filtering sentences.}
\end{table*}

The first attempts we made to directly adjust BART for long text either expanded its positional encodings or segmented the input text and dealt with each segment individually. Below are three ways we attempted.

\paragraph{Arc-I: Position Encoding Expanded BART}
Since the original BART model is pretrained on 512 sequence length and fintuned on 1024 sequence length.\footnote{\url{https://github.com/pytorch/fairseq/issues/1413}} We followed this approach and tried copying the first 1024 position encodings periodically for longer sequence and finetuned the model on our own dataset.

\paragraph{Arc-II: Independently-windowed BART}
In this architecture, we simply chunked the documents into multiple windows with 1024 window size, and then use BART to encode them separately. The final output of the encoder side is the concatenation of those window outputs. The decoder can then generate texts as normal while attending to the whole input representations.

\paragraph{Arc-III: Dependently-windowed BART} In \textbf{Arc-II}, we ignore the interdependence between each chunk which may lead to incoherence in generated texts. Here, to model the inter-window dependencies, we followed the approach introduced in \citet{Rae2020CompressiveTF}. We kept a compressive memory of the past and used this memory to compute the representation of new window. The final output of the encoder side is the concatenation of those window outputs as in \textbf{Arc-II}. \\

However, we found that none of these adjustments could generate satisfying fluent and coherent texts according to our experiments. Common problems include interchanges between first and third person narration (They... Our model...), contradiction between consecutive sentences, more descriptive texts and fewer opinions, etc.

\subsection{CE Extraction Details}
\label{appendix:cem}

The basic sentence statistics of our \texttt{ASAP-Review} dataset is listed in Tab.~\ref{table4}.
\begin{table}[h]
\small
    \centering
    \begin{tabular*}{0.43\textwidth}{@{}lccc@{}}
\toprule
                            & ICLR & NeurIPS & Both \\ \midrule
\hspace{5mm}Avg. Sentence Num.          & 216  & 198  & 209 \\  \bottomrule
\end{tabular*}
    \caption{Sentence statistics of \texttt{ASAPReview} dataset. ``Avg. Sentence Num.'' denotes average sentence number in a paper.}
    \label{table4}
\end{table}

We use two steps to extract salient sentences from a source document: (i) Keywords filtering, (ii) Cross-entropy method

\subsubsection{Keywords Filtering}
We have predefined 48 keywords and in the first stage, we select sentences containing those keywords as well as their inflections.
The 48 keywords are shown in Tab.~\ref{table: keywords}. After applying keywords filtering, the statistics of selected sentences are shown in Tab.~\ref{table5}.

\begin{table}[h]
\small
    \centering
    \begin{tabular*}{0.43\textwidth}{@{}lccc@{}}
\toprule
                            & ICLR & NeurIPS & Both \\ \midrule
\hspace{5mm}Avg. Sentence Num.          & 97  & 85  & 92  \\  \bottomrule
\end{tabular*}
    \caption{Sentence statistics of selected sentences after keywords filtering. ``Avg. Sentence Num.'' denotes average selected sentence number in a paper.}
    \label{table5}
\end{table}

\subsubsection{Cross Entropy Method}

Following \citet{10.1145/3077136.3080690}'s approach in unsupervised summaization. We formalize the sentence extraction problem as a combinatorial optimization problem. Specifically, we define the performance function $R$ as below.
\begin{align}
    R(S) &= -\sum_{w\in S}p_S(w)\log p_S(w)\\
    p_S(w) &= \frac{\text{Count}(w)}{\text{Len}(S)}
\end{align}

Where $S$ represents the concatenation of selected sentences, $\text{Len}(S)$ represents the number of words in $S$ while $\text{Count}(w)$ represents the number of times $w$ appears in $S$. The intuition behind this performance function is that we want to select sentences that can cover more diverse words. Note that when calculating $R(S)$, we do preprocessing steps (i.e. lowercasing, removing punctuation, removing stop words etc.).

For each paper containing $n$ sentences, we aim to find a binary vector $p=(p_1, \cdots, p_n)$ in which $p_i$ indicates whether the $i$-th sentence is selected such that the conbination of selected sentences achieves highest performance score and also contains fewer than 30\footnote{This number is chosen according to our empirical observations. We need to extract sentences that can fit BART's input length (1024).} sentences. We did this by using Cross Entropy Method \cite{rubinstein2013cross}. The algorithm is shown below.

\begin{enumerate}
    \item For each paper containing $n$ sentences, we first assume that each sentence is equally likely to be selected. We start with $p_0 = (1/2, 1/2, ..., 1/2)$. Let $t:= 1$.
    \item Draw a sample $X_1, \cdots, X_N$ of Bernoulli vectors with success probability vector $p_{t-1}$. For each vector, concatenate the sentences selected and get $N$ sequences $S_1, \cdots, S_N$. Calculate the performance scores $R(S_i)$ for all $i$, and order them from smallest to biggest, $R_{(1)} \leq R_{(2)} \leq \cdots \leq R_{(N)}$. Let $\gamma_t$ be $(1 - \rho)$ sample quantile of the performances:
    $\gamma_t = R_{(\lceil (1-\rho)N \rceil)}$.
    \item Use the same sample to calculate
    $\hat{p_t} = (\hat{p}_{t,1}, \cdots, \hat{p}_{t,n})$ via
    \begin{equation}
        \hat{p}_{t,j} = \frac{\sum_{i=1}^N I_{\{R(S_i) \geq \gamma_t\}}I_{\{X_{ij}=1\}}}{\sum_{i=1}^N I_{\{R(S_i) \geq \gamma_t\}}}
    \end{equation}
    where $I_{\{c\}}$ takes the value 1 if $c$ is satisfied, otherwise 0.
    \item Perform a smoothed update.
    \begin{equation}
        p_t = \alpha \hat{p_t} + (1 - \alpha) p_{t-1}
    \end{equation}
    \item If the value of $\gamma_t$ hasn't changed for 3 iterations, then stop. Otherwise, set $t:=t+1$ and return to step 2.
\end{enumerate}

The elements in $p_t$ will converge to either very close to 0 or very close to 1. And we can sample from the converged $p_t$ to get our extraction.

We chose $N=1000$, $\rho=0.05$ and $\alpha=0.7$ when we ran this algorithm. If we happen to select more than 30 sentences in a sample, we drop this sample. Note that we slightly decrease the initial probability when there are more than 90 sentences after filtering to ensure enough sample number in the first few iterations.

\subsection{Detailed Analysis and Case Study}
\label{app:detail_result_analysis}

We take our aspect-enhanced model using CE extraction to conduct case study.
Tab.~\ref{tab:example-aspect} lists five examples for each aspect the model mentions. It can be seen that the language use of generated reviews are pretty close to real reviewers.

\paragraph{Evidence-sensitive}
For aspect-enhanced model, It would also be interesting to trace back to the evidence when the model generates a specific aspect. To do that we inspect where the model attends when it generates a specific aspect by looking at the attention values with respect to the source input.\footnote{The way we aggregate attention values is to take the maximum, no matter is to aggregate tokens to a word or to aggregate different attention heads or to aggregate words to an aspect.}

\begin{figure*}
    \centering
    \includegraphics[width=1\linewidth]{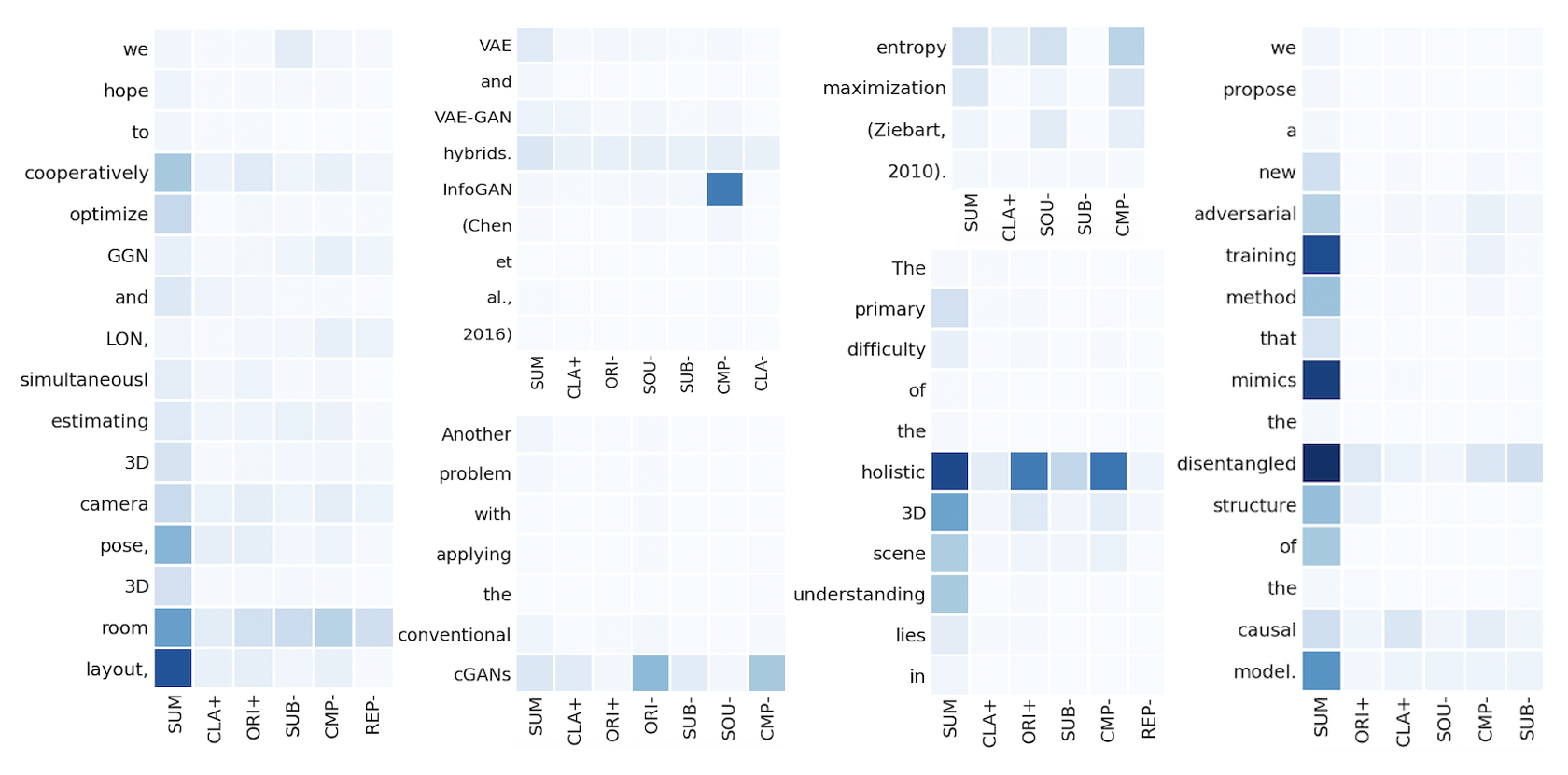}
    \caption{Attention heatmap between source document and generated reviews. $+$ denotes positive sentiment and $-$ denotes negative sentiment.}
    \label{fig:segment_attn}
\end{figure*}

And interestingly, we found that the model attends to the reasonable place when it generates a specific aspect. Fig.~\ref{fig:segment_attn} presents the attention heatmap of several segment texts, the bottom of the figure shows aspects the model generates.
There are some common patterns we found when we examined the attention values between the source input and output.
\begin{enumerate}
    \item When the model generates summary, it will attend to sentences that contain strong indicators like ``\texttt{we propose}'' or ``\texttt{we introduce}''.
    \item When it generates originality, it will attend to previous work part as well as places describing contributions of this work.
    \item When it generates substance, it will attend to experiment settings and number of experiments conducted;
    \item When it generates meaningful comparison, it will attend to places contains ``\texttt{et al.}''
\end{enumerate}

\subsection{Calculation of Aspect Score}
\label{app:aspect_score}
For accepted (rejected) papers, we calculate the average aspect score for each aspect.

The aspect score of a review is calculated as follows.
\begin{itemize}
    \item If an aspect does not appear in a review, then we count the score for this aspect as 0.5 (which stands for neutral)
    \item If an aspect appears in a review, we denote its occurrences as $\mathcal{O} = \{o_1, o_2, \cdots, o_n\}$ where $n$ is the total number of occurrences. And we denote the positive occurrences of this aspect as $\mathcal{O}_p = \{o_{p_1}, o_{p_2}, \cdots, o_{p_n}\}$ where $p_n$ is the total number of positive occurrences.
    The aspect score is calculated using Formula \ref{eq:aspect-score}.
    \begin{equation}
    \label{eq:aspect-score}
        \text{Aspect Score} = \frac{p_n}{n}
    \end{equation}
\end{itemize}

\subsection{Bias Analysis for All Models}
\label{app: bias-analysis}
\begin{figure*}[!htbp]
    \centering
    \includegraphics[width=0.95\linewidth]{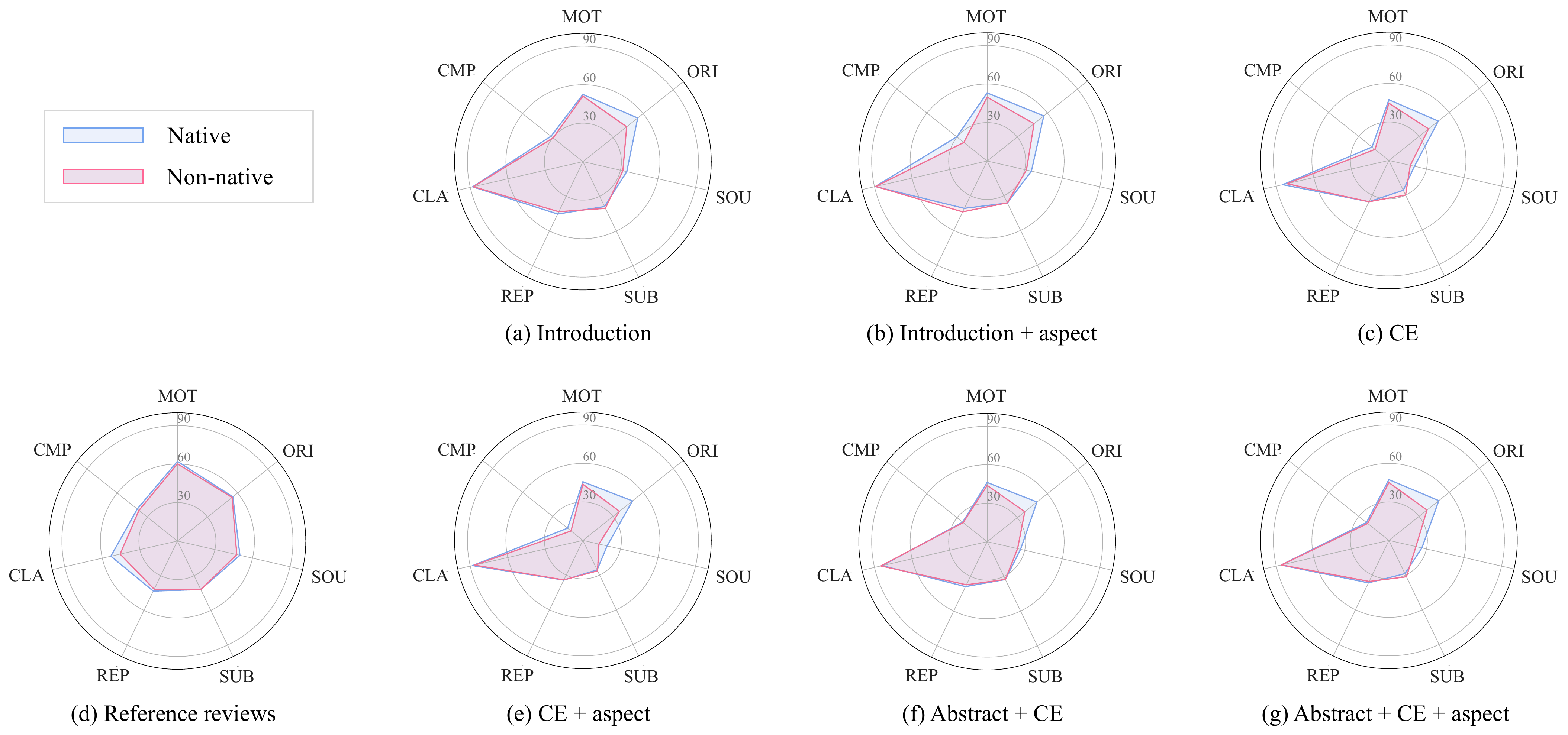}
    \caption{Spider chart of aspect scores for all models with regard to nativeness.}
    \label{fig:all-native-bias}
\end{figure*}

\begin{figure*}[!htbp]
    \centering
    \includegraphics[width=0.95\linewidth]{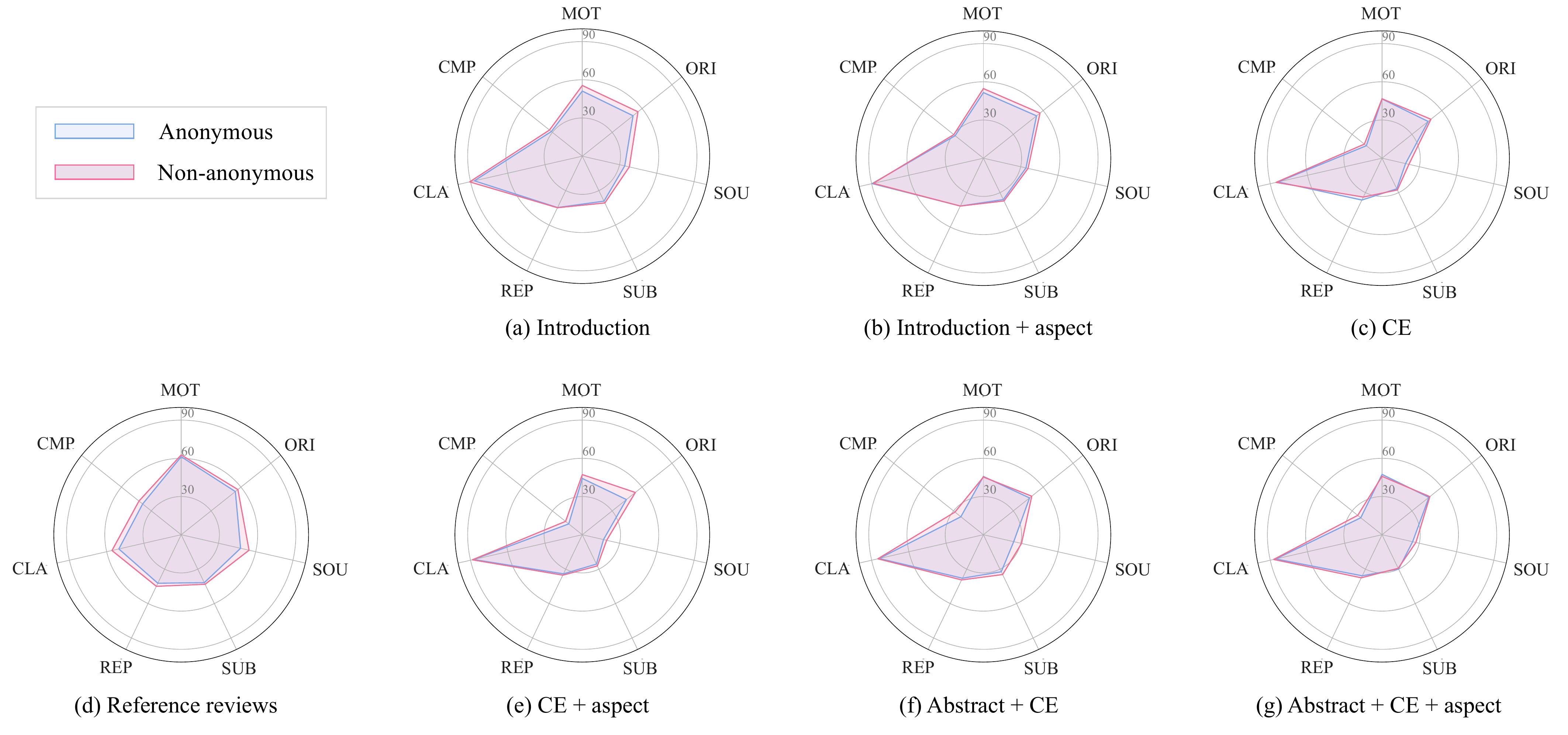}
    \caption{Spider chart of aspect scores for all models with regard to anonymity.}
    \label{fig:all-anonymity-bias}
\end{figure*}

Here, following the methods we proposed in \S\ref{sec:measure_bias}, we list the bias analysis for all models in Fig.~\ref{fig:all-native-bias}, Fig.~\ref{fig:all-anonymity-bias}, Tab.~\ref{tab:all-native-bias}, Tab.~\ref{tab:all-anonymity-bias}.

\begin{table*}[htp]
\centering
\setlength\tabcolsep{10pt}
\small
\renewcommand{\arraystretch}{1.2}
\begin{tabular*}{\textwidth}{@{}lcccccccc@{}}

\toprule
 & MOT & ORI & SOU & SUB & REP & CLA & CMP & Total\\ \midrule
INTRO & -0.72 & +18.71 & +3.84 & -3.66 & +0.73 & -13.32 & +2.40 & 43.39
\\
INTRO+ASPECT & +3.12 & +15.75 & +6.14 & +0.66 & -10.61 & -13.50 & +19.05 & 68.84\\
CE & +2.56 & +18.33 & +11.16 & -13.41 & -3.71 & -9.94 & +13.49 & 72.58 \\
CE+ASPECT & +1.13 & +24.77 & +28.78 & -2.92 & -3.18 & -12.02 & +18.36 & 91.18 \\
ABSTRACT+CE & +1.77 & +23.01 & +3.79 & +0.44 & +0.37 & -15.18 & -2.13 & 46.69 \\
ABSTRACT+CE+ASPECT & +1.72 & +22.23 & +12.94 & -8.30 & -0.38 & -13.40 & +0.89 & 59.86 \\

\bottomrule
\end{tabular*}

\caption{\label{tab:all-native-bias} Disparity differences regarding nativeness. Total is the sum of absolute value of disparity difference.}

\end{table*}

\begin{table*}[htp]
\centering
\setlength\tabcolsep{10pt}
\small
\renewcommand{\arraystretch}{1.2}
\begin{tabular*}{\textwidth}{@{}lcccccccc@{}}

\toprule
  & MOT & ORI & SOU & SUB & REP & CLA & CMP & Total\\ \midrule
INTRO & -5.69 & -4.43 & +2.76 & -0.64 & +5.65 & +5.80 & +3.02 & 28.00 \\
INTRO + ASPECT & -3.53 & -1.65 & +7.85 & +0.01 & +5.93 & +11.02 & +4.20 & 34.20 \\
CE & +1.89 & -1.18 & +0.05 & -0.44 & +13.09 & +8.00 & -2.56 & 27.21 \\
CE+ASPECT & -4.20 & -12.32 & -0.52 & -2.57 & +2.70 & +8.75 & -10.31 & 41.37 \\
ABSTRACT+CE & +3.18 & -0.05 & -7.96 & -3.73 & +2.25 & +8.69 & -12.02 & 37.88 \\
ABSTRACT+CE+ASPECT & +5.45 & +2.49 & +2.80 & +5.69 & +1.33 & +8.03 & -3.79 & 29.59 \\

\bottomrule
\end{tabular*}

\caption{\label{tab:all-anonymity-bias}Disparity differences regarding anonymity. Total is the sum of absolute value of disparity difference.}
\end{table*}

\begin{table*}[h]
\renewcommand{\arraystretch}{1.1}
\small
    \centering
    \begin{tabular*}{0.9\textwidth}{l}
\toprule
\hspace{3mm}\textbf{Motivation} \\ \midrule
\hspace{3mm}1. The motivation of using the conditional prior is unclear.\\
\hspace{3mm}2. I think this paper will be of interest to the NeurIPS community.\\
\hspace{3mm}3. The idea of continual learning is interesting and the method is well motivated. \\
\hspace{3mm}4. Overall, I think this paper is a good contribution to the field of adversarial robustness.\\
\hspace{3mm}5. It is hard to understand the motivation of the paper and the motivation behind the proposed methods.\\ \midrule

\hspace{3mm}\textbf{Originality} \\ \midrule
\hspace{3mm}1. This paper presents a novel approach to cross-lingual language model learning.\\
\hspace{3mm}2. The novelty of the paper is limited . The idea of using low rank matrices is not new.\\
\hspace{3mm}3. The proposed method seems to be very similar to the method of Dong et al. ( 2018 ).\\
\hspace{3mm}4. The idea of using neural networks to learn edit representations is interesting and novel .\\
\hspace{3mm}5. The proposed method seems to be a simple extension of the batched-E-step method proposed by Shazeer \\
\hspace{7mm}et al.\\\midrule

\hspace{3mm}\textbf{Soundness} \\ \midrule
\hspace{3mm}1. This assumption is not true in practice .\\
\hspace{3mm}2. The experimental results are not very convincing .\\
\hspace{3mm}3. But the authors do not provide any theoretical justification for this claim.\\
\hspace{3mm}4. The theoretical results are sound and the experimental results are convincing.\\
\hspace{3mm}5. The paper does not provide any insights on the reasons for the success of the supervised methods.\\ \midrule

\hspace{3mm}\textbf{Substance} \\ \midrule
\hspace{3mm}1. The experiments are well-conducted.\\
\hspace{3mm}2. The ablation study in Section A.1.1 is not sufficient.\\
\hspace{3mm}3. It would be better to show the performance on a larger dataset.\\
\hspace{3mm}4. The authors should show the performance on more difficult problems.\\
\hspace{3mm}5. The experiments are extensive and show the effectiveness of the proposed method.\\\midrule

\hspace{3mm}\textbf{Replicability} \\ \midrule
\hspace{3mm}1. It is not clear how the network is trained.\\
\hspace{3mm}2. The authors should provide more details about the experiments.\\
\hspace{3mm}3. The authors should provide more details about the hyperparameters.\\
\hspace{3mm}4. The authors should provide more details about the training procedure.\\
\hspace{3mm}5. It would be better if the authors can provide more details about the hyperparameters of LST.\\
 \midrule

\hspace{3mm}\textbf{Meaningful Comparison} \\ \midrule
\hspace{3mm}1. The author should compare with [ 1 , 2 , 3 ] and [ 4 ] .\\
\hspace{3mm}2. The authors should compare the proposed method with existing methods .\\
\hspace{3mm}3. It would be more convincing if the authors can compare with other methods such as AdaGrad.\\
\hspace{3mm}4. authors should compare the performance with the state-of-the-art methods in real-world applications .\\
\hspace{3mm}5. I also think the paper should compare the performance of intrinsic fear with the other methods proposed \\
\hspace{7mm}in [ 1 , 2 , 3 , 4 , 5 ].\\\midrule

\hspace{3mm}\textbf{Clarity} \\ \midrule
\hspace{3mm}1. There are some typos in the paper.\\
\hspace{3mm}2. The paper is well-written and easy to follow.\\
\hspace{3mm}3. It is not clear to me how to interpret the results in Table 1.\\
\hspace{3mm}4. It would be better if the authors can provide a more detailed explanation of the difference.\\
\hspace{3mm}5. The paper is not well organized . It is hard to follow the description of the proposed method.\\
\bottomrule
\end{tabular*}
    \caption{Examples for different aspect mention from generated reviews.}
    \label{tab:example-aspect}
\end{table*}

\section{Supplemental Material}
\label{sec:supplemental}

\subsection{Dataset Annotation Guideline}
\label{supp:data-annotation}
The annotation guideline for annotating aspects in reviews can be found at \url{https://github.com/neulab/ReviewAdvisor/blob/main/materials/AnnotationGuideline.pdf}

\end{document}